\documentclass[10pt,twocolumn,letterpaper]{article}

\usepackage[pagenumbers]{cvpr} 

\usepackage{graphicx}
\usepackage{amsmath}
\usepackage{amssymb}
\usepackage{booktabs}
\usepackage{times}  
\usepackage{helvet}  
\usepackage{courier}  
\usepackage[hyphens]{url}  
\usepackage{makecell}
\usepackage{algorithm}
\usepackage{algorithmic}
\usepackage{newfloat}
\usepackage{listings}
\usepackage{amsfonts,pifont,multirow}
\usepackage{xcolor,colortbl}
\definecolor{LightCyan}{RGB}{231, 254, 255}
\definecolor{Yellow}{RGB}{255, 249, 196}
\definecolor{Green}{RGB}{241, 248, 233}
\definecolor{Red}{RGB}{252, 228, 236}
\definecolor{Blue}{RGB}{0, 0,  0}

\usepackage[pagebackref,breaklinks,colorlinks]{hyperref}

\newcommand{\ve}[1]{\mathbf{#1}}
\newcommand{\vv}[1]{\mbox{\boldmath $#1$}}
\newcommand{\head}[1]{{\noindent\textbf{#1}}}
\usepackage[capitalize]{cleveref}
\crefname{section}{Sec.}{Secs.}
\Crefname{section}{Section}{Sections}
\Crefname{table}{Table}{Tables}
\crefname{table}{Tab.}{Tabs.}

\def\cvprPaperID{*****} 
\def\confName{CVPR}
\def\confYear{2022}
\makeatletter

\begin{document}

\title{Symbolic Replay: Scene Graph as Prompt for Continual Learning on VQA Task}
\author{
Stan Weixian Lei\textsuperscript{1}$^*$, Difei Gao\textsuperscript{1}$^*$, Jay Zhangjie Wu\textsuperscript{1}, Yuxuan Wang\textsuperscript{1},\\ Wei Liu\textsuperscript{2}, Mengmi Zhang\textsuperscript{3}, Mike Zheng Shou\textsuperscript{1}$^{\dag}$\\
 \\
\textsuperscript{1}Show Lab, National University of Singapore \\
\textsuperscript{2}Tencent Data Platform \\
\textsuperscript{3}CFAR and I2R, Agency for Science, Technology and Research
}

\maketitle
\def\thefootnote{*}\footnotetext{Equal contribution. $^{\dag}$Corresponding author.}\def\thefootnote{\arabic{footnote}}


\begin{abstract}
VQA is an ambitious task aiming to answer any image-related question. However, in reality, it is hard to build such a system once for all since the needs of users are continuously updated, and the system has to implement new functions. 
Thus, Continual Learning (CL) ability is a must in developing advanced VQA systems. 
Recently, a pioneer work split a VQA dataset into disjoint answer sets to study this topic. However, CL on VQA involves not only the expansion of label sets (new \textbf{A}nswer sets). It is crucial to study how to answer questions when deploying VQA systems to new environments (new \textbf{V}isual scenes) and how to answer questions requiring new functions (new \textbf{Q}uestion types).
Thus, we propose CLOVE, a benchmark for \textbf{C}ontinual \textbf{L}earning \textbf{O}n \textbf{V}isual qu\textbf{E}stion answering, which contains scene- and function-incremental settings for the two aforementioned CL scenarios. 
In terms of methodology, the main difference between CL on VQA and classification is that the former additionally involves expanding and preventing forgetting of reasoning mechanisms, while the latter focusing on class representation. 
Thus, we propose a real-data-free replay-based method tailored for CL on VQA, named Scene Graph as Prompt for Symbolic Replay. Using a piece of scene graph as a prompt, it replays pseudo scene graphs to represent the past images, along with correlated QA pairs. 
A unified VQA model is also proposed to utilize the current and replayed data to enhance its QA ability.
Finally, experimental results reveal challenges in CLOVE and demonstrate the effectiveness of our method. 
The dataset and code will be available at~\url{https://github.com/showlab/CLVQA}.
\end{abstract}

\section{Introduction}
\label{sec:intro}

\begin{figure}[htb]
\centering
\includegraphics[width=\columnwidth]{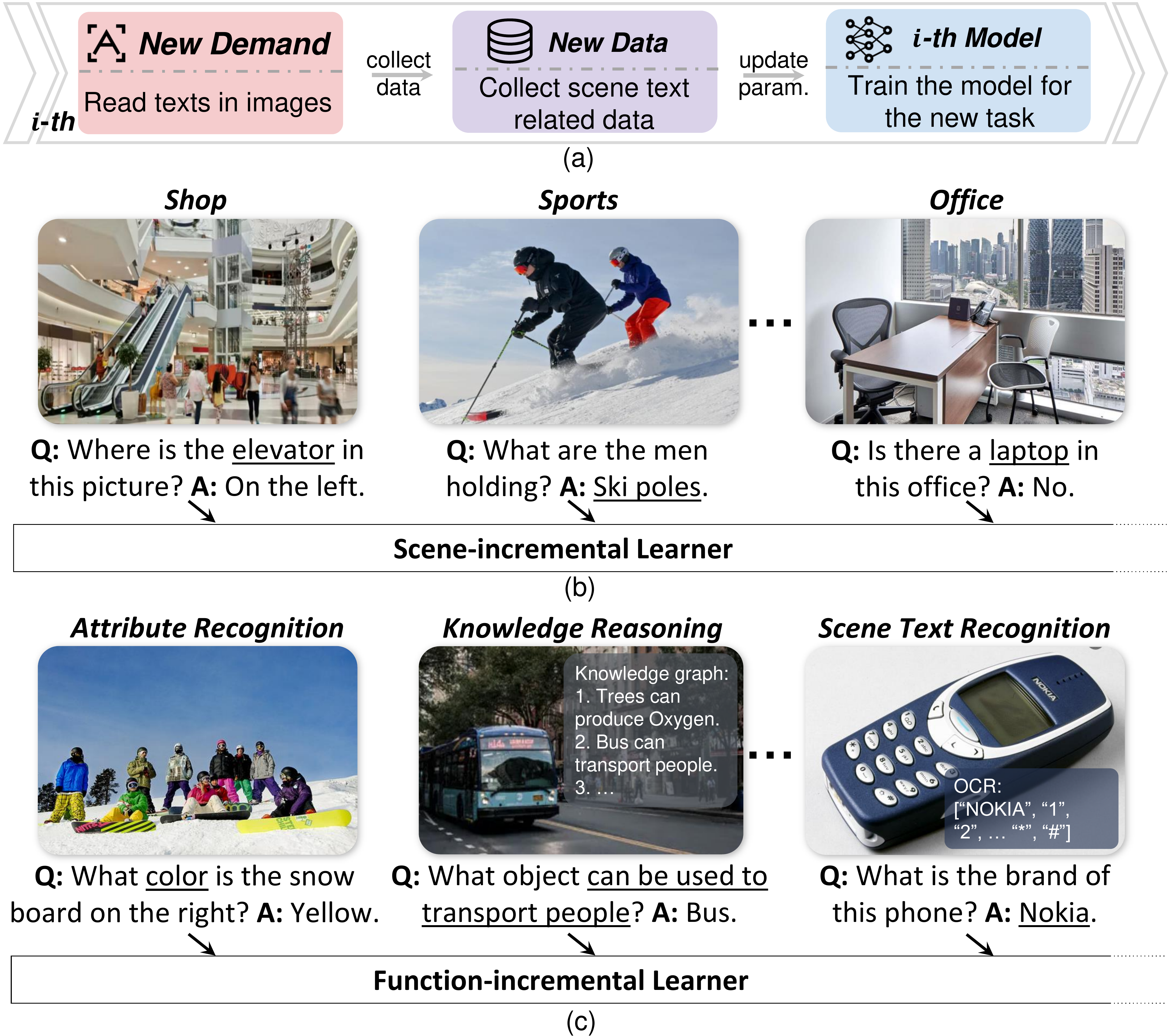}
\caption{
(a) An AI continuously receives new demands and updates itself with collected data. 
(b) A scene-incremental learner adapts to new scenes for deployment. 
(c) A function-incremental learner acquires new functions over time.
}
\label{fig:teaser}
\end{figure}

In recent years, we have witnessed tremendous successes in achieving state-of-the-art performance on VQA tasks by CV and NLP communities~\cite{anderson2018bottom,lu2019vilbert, su2019vl,jiang2020defense,chen2020uniter}. 
Despite the remarkable success, current VQA systems are usually trained on specific datasets and then fixed for use. However, in real applications, user's demands are always updated, 
as is shown in~\cref{fig:teaser}(a). The VQA system is expected to continuously learn the knowledge, when it is deployed to new scenes or need to add new functions. 

In the CV community, various approaches have been studied actively to tackle continual learning for image classification~\cite{rebuffi2017icarl,li2017lwf}, where a model is trained sequentially on a set of images with disjoint labels. 
In this setting, a model continually enhances one ability (\textit{i.e.}, recognition) over one modality (\textit{i.e.}, vision) and mainly learns the representation for each class.
In VQA, a model is required to adapt to new environments (\textit{e.g.}, shop, office, etc.) or learn new abilities (\textit{e.g.}, attribution recognition, knowledge reasoning, etc.) according to the changing demands, as shown in Fig~\ref{fig:teaser} (b) and (c). 
As such, CL in VQA is different from the aforementioned image classification that a model continually learns new ablities over multi-modalities and focuses more on the reasoning.
Thus, it is crucial to study Continual Learning for Visual Question Answering (CLVQA).

 Only a few pioneer works attempted to explore CLVQA. ~\cite{greco-etal-2019-psycholinguistics} studied the continual learning on two subsets, wh- and yes/no questions, of synthesized CLEVR dataset~\cite{Johnson_2017_CVPR_clevr} split by themselves.
However, using the above setting is still hard to evaluate the continual learning of new scenes or functions we are interested in. Specifically, for the image domain, images for both wh- and yes-no questions are about geometric spheres. For functions, both types of questions are about the object properties and multi-hop reasoning upon them. The only difference is the way to ask, \textit{e.g.}, \textit{what color is the cube?}, \textit{is the cube in red?} actually evaluating the same function set.

Thus, we reorganize existing VQA datasets~\cite{hudson2019gqa, gao2019two, singh2019towards} to construct CLOVE, a novel benchmark devised for \textbf{C}ontinual \textbf{L}earning \textbf{O}n \textbf{V}isual qu\textbf{E}stion answering with two continual learning settings, as shown in~\cref{fig:teaser}(b) and~\cref{fig:teaser}(c).
(1) \textit{Scene-incremental setting} mimics the scenario where a VQA agent is adapted to new scenes for deployment. Our CLOVE-scene contains 6 scenes: \textit{ShopAndDinning}, \textit{Workplace}, \textit{HomeOrHotel}, \textit{Transportation}, \textit{SportAndLeisure} and \textit{Outdoors}.
(2) \textit{Function-incremental setting} tests a model's ability in acquiring new functional abilities over time. Our CLOVE-function contains 6 functions: \textit{object recognition}, \textit{attribute recognition}, \textit{relation reasoning}, \textit{logic reasoning}, \textit{knowledge reasoning} and \textit{scene text understanding}.

In terms of methodology, traditional continual learning methods on image classification ~\cite{rebuffi2017icarl} are specially designed to prevent forgetting the representation of vision modality. 
They could be hard to perform well on CLVQA, which requires multi-modal reasoning. Specifically, regularization-based methods~\cite{kirkpatrick2017ewc,aljundi2018mas} might fail in estimating the importance and balancing the learning of past and new tasks due to the complicated model design in VQA models.
Some other works preserve historical knowledge through the replay. And since, in the CL setting, real data usually cannot be saved due to privacy concerns, many works retain knowledge by generating pseudo samples~\cite{shin2017gen_replay}. However, replaying pseudo samples in CLVQA could be extremely challenging. The images could come with complicated visual scenes and fine-grained details, which could be hard to be precisely generated by the state-of-the-art image GAN model~\cite{sauer2022stylegan}. Generated images in low quality also limits the quality of generated question-answer pairs. 
All these issues pose a dilemma for generating \textit{image-question-answer} for pseudo-replay.

In this paper, we introduce Scene Graph as Prompt for symbolic replay (SGP), a real-data-free replay-based method for CLVQA. SGP overcomes the aforementioned limitations of replay based methods by leveraging the scene graph, a concise and structured representation of visual information, as an alternative to images for replay. 
Specifically, SGP consists of a symbolic replay model (SRM) and a unified VQA model (UniVQA). The SRM, which belongs to a language model~\cite{radford2019gpt2}, continuously captures symbolic reasoning mechanism and learns the task-specific mapping between scene graph and QA pairs.
During inference, SRM replays the \textit{scene-graph-question-answer} triplet for knowledge revisiting, prompted by a randomly sampled scene graph relationship. We call this ``symbolic replay''.
Besides, the UniVQA is designed to adapt a wide range of input modalities for different VQA tasks. Trained with the mix of current task samples and symbolic-replayed samples, UniVQA is capable of learning a new task while retaining the previously acquired knowledge. Moreover, since the past real data is not saved, our framework can be employed to various situations involving privacy concerns. 

Extensive experiments with various types of existing CL methods and our model show the difficulties of our benchmark and demonstrate the effectiveness of our method.

\section{Related Work}
\label{sec:related_work}
\subsection{Visual Question Answering.}
VQA is a general task aiming to answer any image-related question. It requires AI to achieve a vast set of functions to answer questions, ranging from fine-grained recognition, object detection, activity recognition to commonsense reasoning, etc.

~\cite{Antol_2015_ICCV_vqa, balanced_vqa_v2} introduced the VQA benchmarks for understanding the common visual concepts in real world.
~\cite{Johnson_2017_CVPR_clevr} built a synthetic dataset for testing visual reasoning abilities, \textit{e.g.}, multi-hop and logic reasoning. 
~\cite{hudson2019gqa} constructed a VQA dataset sourced from Visual Genome~\cite{krishna2017visual_genome} by leveraging the annotated scene graph, aiming to test the model’s compositional reasoning capability in real images. ~\cite{gao2019two,marino2019ok} proposed benchmarks where a model should resort to external knowledge for reasoning. ~\cite{singh2019towards} built a dataset where a model should understand the text in images to answer questions. 

The development of VQA shows that the function set of VQA always need to be continuously expanded with demand.
However, few benchmarks focus on continual learning on VQA.
Thus, we propose a benchmark to mirror real-world scenarios where an AI is required to be deployed to new environments or learn new functions in a CL manner.  

\subsection{Continual Learning Benchmarks.}
In CV community, most of works study continual learning under three settings: 
(1) Class-incremental learning, where a classification model learns to classify increasing number of classes over time~\cite{li2017lwf,rebuffi2017icarl}.
(2) Task-incremental learning, where task identity of newly included task (a set of classes) remains known during inference~\cite{aljundi2017expert,serra2018overcoming}.
(3) Domain-incremental learning, where a model sequentially learns to solve tasks with shifts in input distributions~\cite{rebuffi2017learning_DIL}. 

In NLP, CL is conducted on tasks with different domains~\cite{ZhiyuanChen2015LifelongLF,SungjinLee2017TowardCL} or on cross-task benchmarks~\cite{biesialska-etal-2020-nlp-survey-continual,hu2020xtreme,sun2019lamol}.

All the aforementioned CL tasks focus on one single modality. And for CV, it focuses only on one single ability, \textit{e.g.}, image classification or detection. In contrast, CLVQA considers multiple modalities, vision and language, and involves mutiple abilities.

Similar to ours, ~\cite{greco-etal-2019-psycholinguistics} proposed a CL benchmark on VQA on the synthesized CLVER dataset.
It selected Wh- and Yes/No-type questions and studied CL in the two yielded task orders.
This setting is similar to class-incremental learning in that the increments are on the disjoint answer set. Unlike CLOVE, neither shift in image domain nor expansion of acquired functions set is reflected. The challenges in CLVQA might be underestimated.

\subsection{Continual Learning Methods.}
Continual learning studies the methods that can learn new knowledge without forgetting the past knowledge. Existing CL methods can be grouped into the following categories: 
(1) Replay-based method, which reminds models of knowledge from previous tasks through experience replay. iCaRL~\cite{rebuffi2017icarl} tackles CIL by selecting the nearest-mean-examples from previous tasks and combining rehearsal and distillation strategies. 
GEM~\cite{lopez2017gradient} preserves a subset of real samples from previous tasks. Utilizing these real samples during optimization helps somewhat constrain parameter gradients.
~\cite{shin2017gen_replay,sun2019lamol} synthesized pseudo samples with generative models to mitigate catastrophic forgetting.
(2) Regularization-based method which tries to keep the weights that are important for the previous tasks. EWC~\cite{kirkpatrick2017ewc} uses the Fisher Information Matrix to estimate the importance while MAS~\cite{aljundi2018mas} estimates by measuring how small changes in the parameters affect the output of the model.
(3) Architecture-based method, where different tasks are associated with different modules of the whole model, reduces the interference between tasks ~\cite{mallya2018packnet,liu2021aanet}.

Most of the above methods mitigate catastrophic forgetting from the perspective of representation. However, VQA tasks inherently involve two types of abilities: representation learning and reasoning. Thus, we propose symbolic replay, which helps continuously learn the representation and reasoning skills over a sequence of multi-modal VQA tasks.

\begin{figure*}[hbt]
\centering
\includegraphics[width=0.90\textwidth]{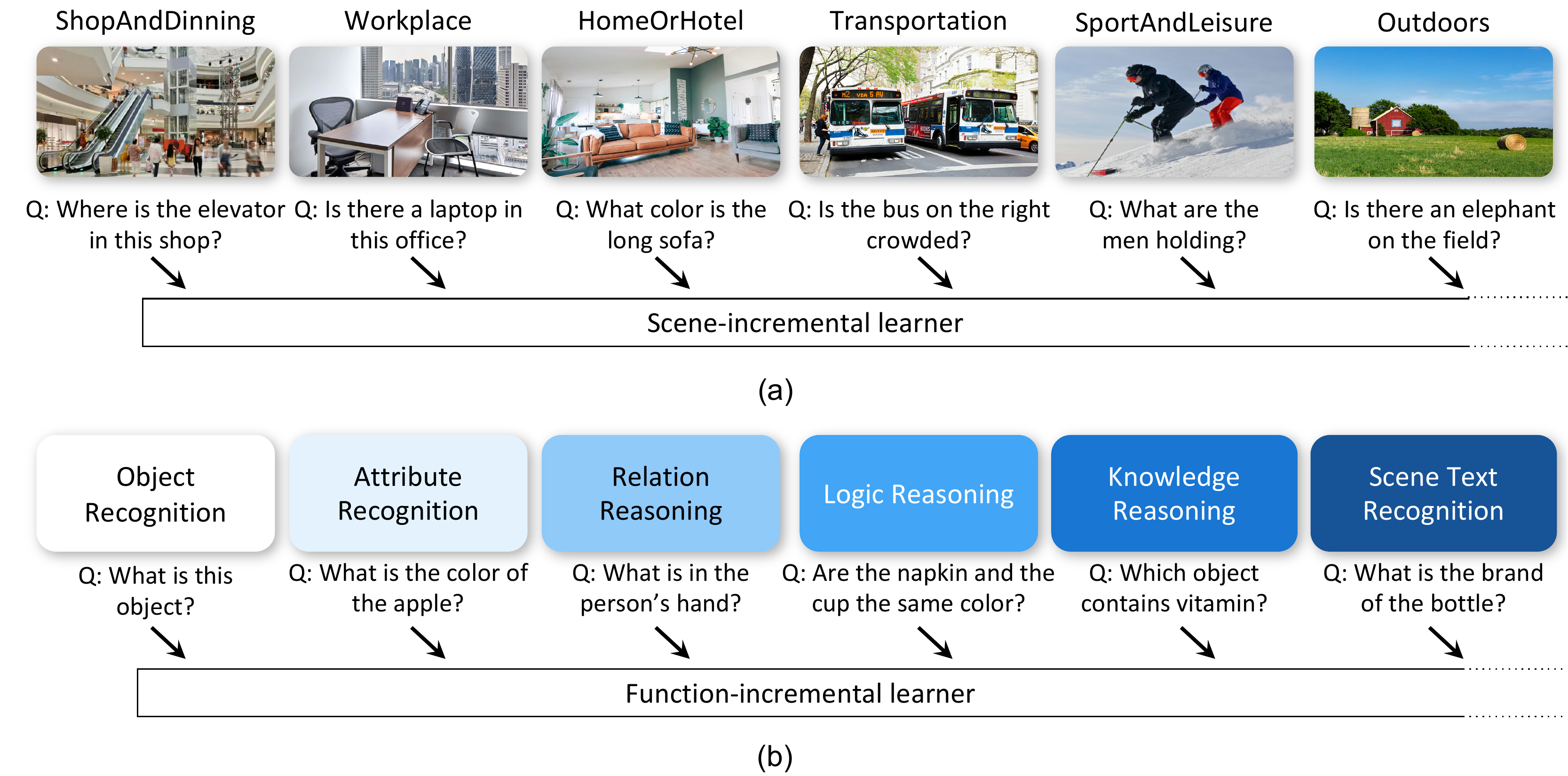} 
\caption{(a) A continual learning paradigm on CLOVE-scene: an agent continuously learns to answer questions when deployed to new scenes: \textit{ShopAndDining} $\rightarrow$ \textit{Workplace} $\rightarrow$ \textit{HomeorHotel} $\rightarrow$ \textit{Transportation} $\rightarrow$ \textit{SportAndLeisure} $\rightarrow$ \textit{Outdoors}. (b)  A continual learning paradigm on CLOVE-function: an agent continuously learns to answer questions to fit the expansion of function set: \textit{object recognition} $\rightarrow$ \textit{attribute recognition }$\rightarrow$ \textit{relation reasoning} $\rightarrow$ \textit{logic reasoning }$\rightarrow$ \textit{knowledge reasoning }$\rightarrow$ \textit{scene text recognition}.
}
\label{fig:supp_benchmark}
\end{figure*}

\section{\includegraphics[scale=0.025, bb=-50 100 550 34]{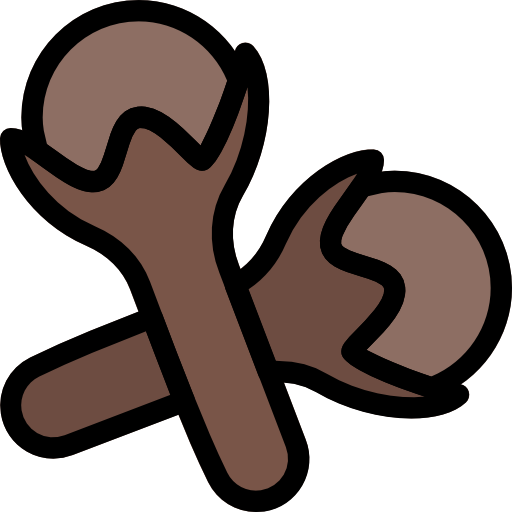}~CLOVE Benchmark}
\label{sec:benchmark}
Here, we detail how we create CLOVE, a benchmark for \textbf{C}ontinual \textbf{L}earning \textbf{O}n \textbf{V}isual qu\textbf{E}stion answering. It contains scene-incremental setting and function-incremental setting, named as CLOVE-scene and CLOVE-function. ~\cref{fig:supp_benchmark} showcases the corresponding continual learning paradigm with examples.

\begin{figure}[ht]
\centering
\includegraphics[width=\columnwidth]{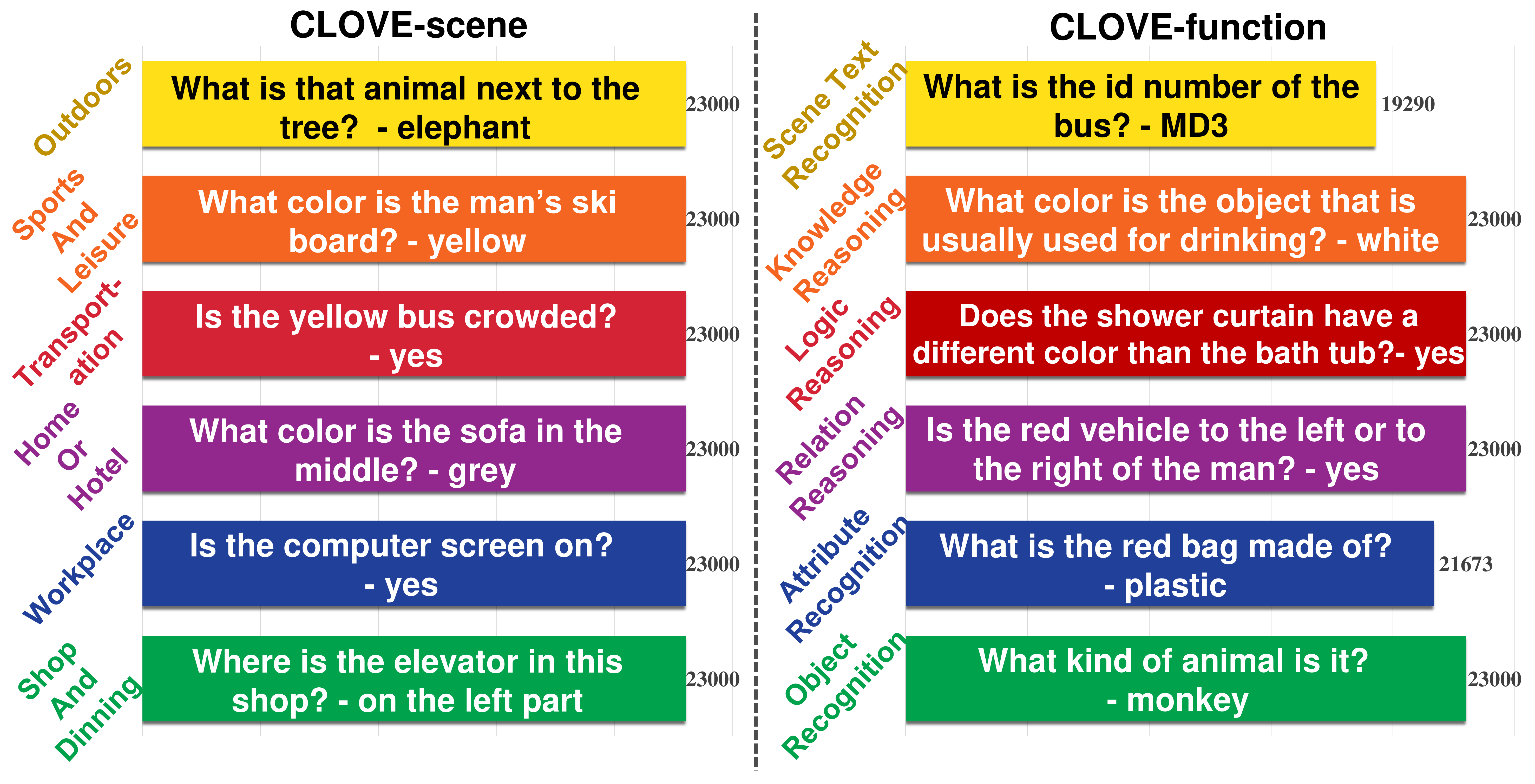} 
\caption{
Sample numbers with one QA example for each task in CLOVE-scene and CLOVE-function. 
}
\label{fig:benchmark}
\end{figure}

\subsection{Task Formulation}
In our VQA continual learning framework, we define the sequence of $N$ VQA tasks to be solved as $\ve{T} =\left(T_1,T_2, \cdots, T_N \right)$, where task $T_i$ is to optimize a model towards an objective on the task-specific dataset $D_i = \{\ve{d}^i_1, \cdots, \ve{d}^i_k, \cdots, \ve{d}^i_{|D_i|}\}$. Here, each $\ve{d}^i_k$ is an image-question-answer triplet $\{ \ve{v}^i_k, \ve{q}^i_k, \ve{a}^i_k \}$. The image and question are inputs to a VQA model, and the ground-truth answer is the desired output.

\subsection{CLOVE-scene Setting}
An advanced AI agent is expected to be capable of answering questions from different scenes. It may need to adapt to a novel visual environment with new concepts, and meanwhile, remember the past knowledge.

\head{Setting definition}. 
We refer to the taxonomy in the SUN database~\cite{xiao2010sun} and classify the sourced images in GQA~\cite{hudson2019gqa}. We obtain six classes from the second level of scene hierarchy defined in the SUN database:
\textit{ShopAndDinning}, \textit{Workplace}, \textit{HomeOrHotel}, \textit{Transportation}, \textit{SportAndLeisure} and \textit{Outdoors}.

\head{Image sourcing.}
To obtain images for each task, we resort to a \textit{sota} scene classification model~\cite{tang2020unbiased} to obtain an initial partition.
Then, we apply two post-processing strategies to improve the qualities of the selected images: (1). filter images with a low classification confidence score; (2) filter images with limit frequent objects in that scene, which is given in the SUN database.
Finally, we randomly sample 100 images from each task and ask 3 human workers to evaluate the accuracy. Result shows that the yielded splitting achieves a mean accuracy of $91.0\%$.

\begin{figure}[htb]
\centering
\includegraphics[width=\columnwidth]{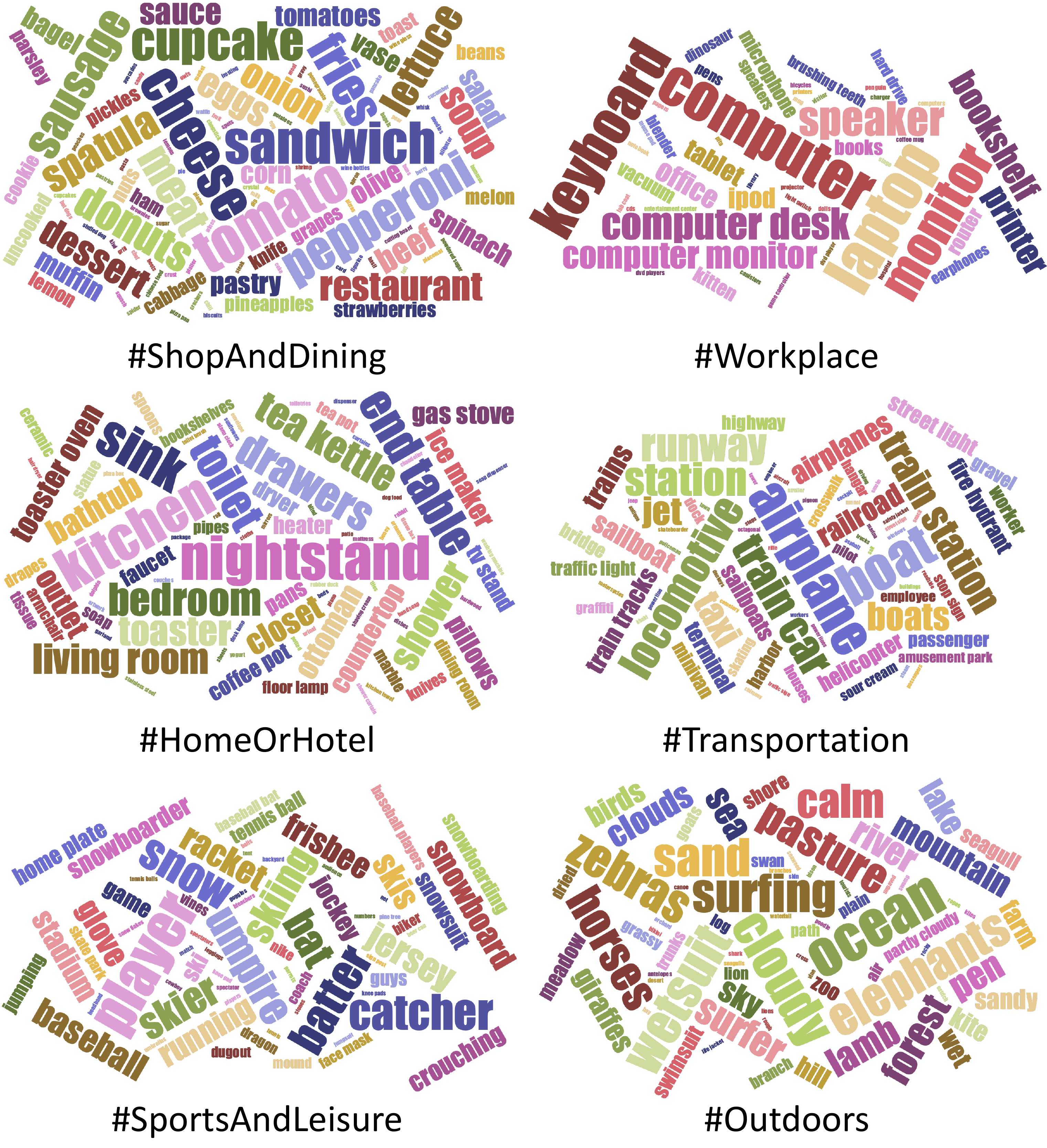} %
\caption{
Unique answers of each task in CLOVE-scene.
}
\label{fig:supp_unique_ans}
\end{figure}

\head{Question-answer pairs sourcing.} 
When deployed to a new scene, the model may face two type of questions:
(1) Questions related to a unique object in that scene, requiring the model to predict the object's name as a scene-specific answer.
(2) Questions related to general concepts shared among different scenes, \textit{e.g.}, color and material, requiring the model to predict common answers.
Our scene incremental setting includes both two types of questions.
Concretely, we maintain a set of common answers for all tasks and sets of unique answers for each task. Specifically, a common answer can appear in different tasks (\textit{e.g.},``red'' can appear in different tasks) while a unique answer is only allowed to appear in its corresponding task (\textit{e.g.}, ``computer monitor'' only appears in \textit{Workplace}). 
For each task, we collect similar number of samples with common answers and with unique answers. Specifically, we showcase the unique answers of each task in CLOVE-scene in~\cref{fig:supp_unique_ans}, from which we can observe the scene-specific feature of the CLOVE benchmark.
In addition, we balance the number of samples among different tasks and follow GQA to smooth the answer distribution within each task to avoid dataset bias. 

\subsection{CLOVE-function Setting}
A VQA system requires a vast set of functions to answer a question – to name a few, object detection (e.g. ``How many birds are there?''); activity recognition (e.g., ``Is the man running?'') , knowledge base reasoning (e.g., ``Is this a vegetarian pizza'') and scene text recognition (``What is the name of this book?''). It is common that an AI agent is required to develop different functions when encountering different use cases over time. To mimic such a scenario, we create CLOVE-function, by sampling and reorganizing data from GQA~\cite{hudson2019gqa}, CRIC~\cite{gao2019two} and TextVQA~\cite{singh2019towards}.

\head{Setting definition.} Based on the functions defined and introduced in GQA, CRIC and TextVQA, we collect six tasks for function incremental setting:
\textit{object recognition}, \textit{attribute recognition}, \textit{relation reasoning}, \textit{logic reasoning}, \textit{knowledge reasoning} and \textit{scene text recognition}. 

\head{Sample sourcing.} We source data from GQA for \textit{object recognition}, \textit{attribute recognition}, \textit{relation reasoning} and \textit{logic reasoning}, 
while directly sourcing samples for \textit{knowledge reasoning} and \textit{scene text recognition} from CRIC and TextVQA, respectively. 
Thanks to the rich annotations provided in GQA and CRIC, for each question we can obtain a functional program that specifies the reasoning steps having to be taken to answer it.
We then define a unique set of function operations for each task. A question with functional program containing a specific operation set is assigned to the corresponding stage, as is shown in~\cref{tab:supp_func_rule}.
Note that the function operations sets of all tasks are not exclusive, as they may share some of the functions which are the bases of VQA.

\noindent\textbf{Distribution smoothing.} To facilitate the study of CLVQA, we create comparable number of samples for each task in order to avoid the potential issues caused by the data imbalanced situation. 

We showcase the statistic information and question examples of CLOVE in Fig~\ref{fig:benchmark}. More details could be found in Supp.

\begin{table}[ht]
\resizebox{0.47\textwidth}{!}{
\begin{tabular}{|c|c|c|}
\hline
\multicolumn{1}{|c|}{Function} & \multicolumn{1}{c|}{Operation}  & \multicolumn{1}{c|}{Argument}          \\ \hline
Object Recognition             & \makecell{\texttt{Select, Query, } \\ \texttt{Choose}}  & \makecell{\texttt{name} }                                 \\ \hline
Attribute Recognition          & \makecell{\texttt{Query, Verify,} \\ \texttt{Choose, Filter}}   & \makecell{\texttt{color, material,} \\ \texttt{weather} $\cdots$}       \\ \hline
Relation Reasoning             & \makecell{\texttt{Relate, Verify} \\ \texttt{Choose}}          & \texttt{rel}                                    \\ \hline
Logic Reasoning                & \makecell{\texttt{Different, Same,} \\ \texttt{Common, Choose}} & \makecell{\texttt{same color,}\\ \texttt{choose healthier,} $\cdots$} \\ \hline
Knowledge Reasoning            & \makecell{\texttt{Find with} \\ \texttt{Knowledge Graph}}       &                  \texttt{ - }                      \\ \hline
Scene Text Recognition         & \makecell{\texttt{scene text} \\ \texttt{recognition}}          &               \texttt{ - }                         \\ \hline
\end{tabular}
}
\caption{Function operations with argument examples. We assign each question to the corresponding task following these rules. }
\label{tab:supp_func_rule}
\end{table}

\subsection{Evaluation Metric}
For each task in CLVQA, we follow VQAv2~\cite{balanced_vqa_v2} to calculate the accuracy for a question, which is measured via soft voting of the 10 answers. The accuracy is defined as $\text{Acc(ans)}=\min\left\{ \frac{\text{\#ans in annotation}}{3}, 1 \right\}$. To measure a model's performance on CLVQA, we use average accuracy. More metrics could be found in Appendix.

\head{Average accuracy.} Let $a_{k,j}$ denote the accuracy evaluated on the held-out testset of $T_j$($j \leq k$) after training a continual learner from $T_1$ to $T_k$. The average accuracy at $T_k$ is defined as $A_k = \frac{1}{k}\sum_{j=1}^{k}a_{k,j}$.

\section{Method}
\label{sec:method}
\begin{figure*}[ht]
\centering
\includegraphics[width=0.95\textwidth]{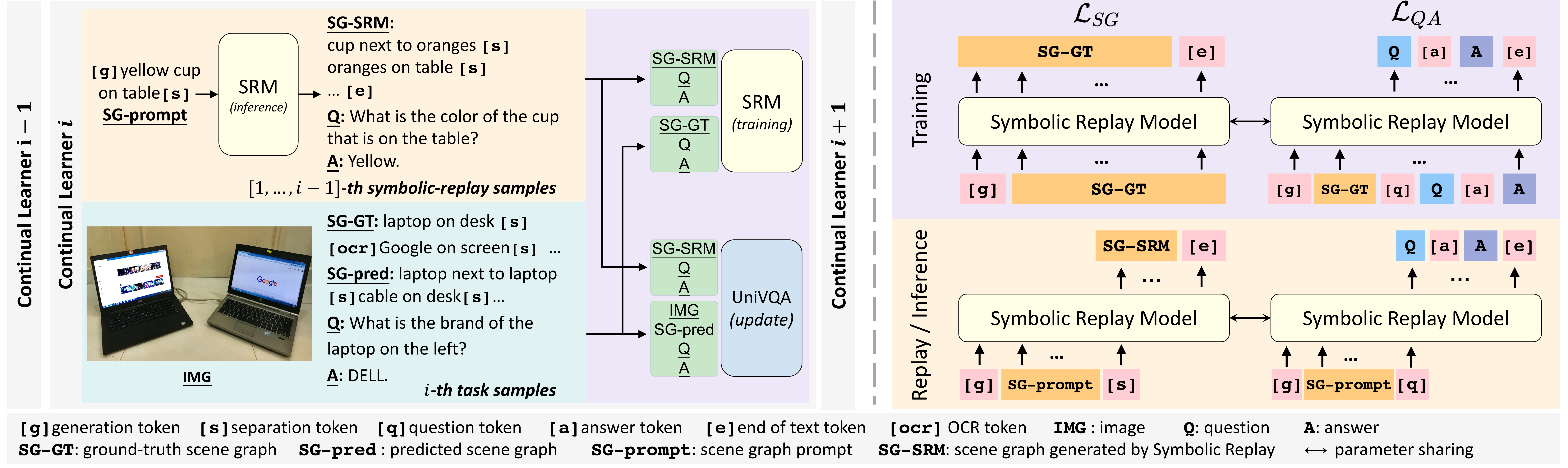}
\caption{
\textbf{Left}: Sequential training of our continual learner. Before training on $T_i$, the SRM takes SG-prompt as input and generates \textit{SG-SRM-question-answer} triplet for replay.
During training, both SRM and UniVQA are trained with the mix of current and replayed samples.
\textbf{Right}: Details of SRM. During training, we apply the next token prediction task on the GT scene graph sequence(SG-GT) and supervise question-answer generation using the related scene graph relationships. During inference, the SRM takes an scene graph relationship (SG-prompt) as input, and outputs the completed scene graph and generated question-answer pair. A detailed example is shown on the \textbf{left} part. 
} 
\label{fig:pipelineNsrm}
\end{figure*}

\subsection{Overview of Continual Learning Pipeline}
In this section, we introduce our SGP framework for CLVQA, which contains a Symbolic Replay Model (SRM), denoted as $S$, and a unified VQA model (UniVQA), denoted as $U$. 
As is shown in~\cref{fig:pipelineNsrm}, training the whole continual learner involves two independent procedures of training $S$ and $U$.
To train $S$, $S$ itself firstly replays scene graph as the representation of $\ve{v}'$ and generates related $\ve{q}'$ and $\ve{a}'$, which is prompted by a random-sampled relationship (SG-prompt).
Combining the annotated scene graph (SG-GT) of $\ve{v}$, $\ve{q}$ and $\ve{a}$ from the current task, 
$S$ learns the potential scene graph representation and QA patterns from the mix of pseudo-replayed and real samples.
To train $U$, we again use the mix of the pseudo-replayed and real samples, enabling $U$ to learn knowledge from both current and previous tasks.

\subsection{Symbolic Replay Model}
\head{Leveraging scene graph.} Given the difficulties in generating an image with complicated scenes and fine-grained details, replaying highly correlated \textit{image-question-answer} triplets in CLVQA could be intractable. Thus, we resort to scene graphs and leverage a language model (our SRM), DistilGPT2~\cite{radford2019gpt2,vaswani2017bert_attention}, for pseudo-replay. Scene graph is a graphical representation for images and is similar to the form widely used in knowledge base representations~\cite{krishna2017visual_genome}. Therefore, it plays a role as the bridge connecting vision and language. In addition, scene graphs can be used to power a question engine to generate diverse questions over an image~\cite{hudson2019gqa}, thus enabling a language model to learn the potential question-answering pattern.

\head{Symbolic replay.} Combining scene graphs and the language model, we propose SRM, a model which uses an SG-prompt for scene graph replay ($\ve{v}'$) and question-answer pair replay ($\ve{q}'$,$\ve{a}'$). We name this symbolic replay.

\head{Scene graph completion.} With the associated scene graph annotations from images, we sequentialize the scene graph for each image and apply next token prediction over the sequence, enabling $S$ to learn the structure of the image. Let $\mathcal{G}=(g_1, g_2, \cdots, g_K)$ denote the scene graph from current or replayed data. During training, we minimize the following objective:
\begin{equation*}
    \mathcal{L}_{SG}(\vv{\theta})= -\sum_{|D|}\sum_{k} \log P(g_k | g_1, \cdots, g_{k-1}; \vv{\theta}),
\end{equation*}
where $P(g_k) =\text{softmax}(S(G_k))$, $G_k$ represents the context tokens of $g_k$, $\vv{\theta}$ is the model parameters of $S$, and $D$ is the training data. We organize the input sequence as shown in~\cref{fig:pipelineNsrm}: the input is the  concatenation of the generation token \texttt{[g]} and the scene graph as input, and the output is obtained by shifting a word of the input sequence and appending an end of text token \texttt{[e]}. Note that within the input scene graph, a separation token \texttt{[s]} is between two individual relationships.
During inference for scene graph replay, we construct the input as a concatenation of \texttt{[g]}, a randomly sampled SG-prompt and \texttt{[s]}, then $S$ decodes SG-SRM in an autoregressive manner to complete the scene graph.

\head{Question-answer generation.} We also adapt $S$ for the supervised question-answer generation task. Let $G_{qa}$ denote the scene graph relationships used to generate the GT question-answer pair. During training, we minimize the following objective:
\begin{equation*}
    \mathcal{L}_{QA}(\vv{\theta}) = -\sum_{|D|} \log P(q,a|G_{qa};\vv{\theta}),
\end{equation*}
where $P(q,a)= \text{softmax}(S(G_{qa}))$. As is shown in ~\cref{fig:pipelineNsrm}, we concatenate \texttt{[g]}, the question-answer related scene graph relations, question token \texttt{[q]}, the question, answer token \texttt{[a]}, and the answer as the input. $S$ learns to decode the sequence of question, \texttt{[a]}, answer and \texttt{[e]}.
During inference for question-answer pair replay, the input to $S$ is the concatenation of \texttt{[g]}, a random-sampled SG-prompt and \texttt{[q]}, and $S$ decodes the question, \texttt{[a]}, answer and \texttt{[e]}.

\head{Joint-training loss.} The loss function for jointly training $S$ is formulated as $\mathcal{L}_{SRM} = \mathcal{L}_{QA} + \lambda \mathcal{L}_{SG}$, where $\lambda$ is the weight of scene graph completion loss.

\head{Use the pseudo-replayed samples.} The current $S$ uses the SG-SRM as input for the next token prediction task and uses the randomly-sampled SG-prompt, generated question and answer to supervise question-answer generation.

\head{Scene-graph prompt for symbolic replay.}  During training $S$, our method does not explicitly save any real scene graphs to external memory, as it may raise privacy issues in real-world applications. Instead, we maintain a scene graph database for each specific task. Concretely, we go through the training set in task $T_i$, and calculate the frequencies of the objects, attributes, relations. Then, we randomly sample one to three scene graph items for replaying based on the statistics, and save them as SG-prompt for further replay.

\subsection{Unified VQA Transformer}
\begin{figure}[ht]
\centering
\includegraphics[width=\columnwidth]{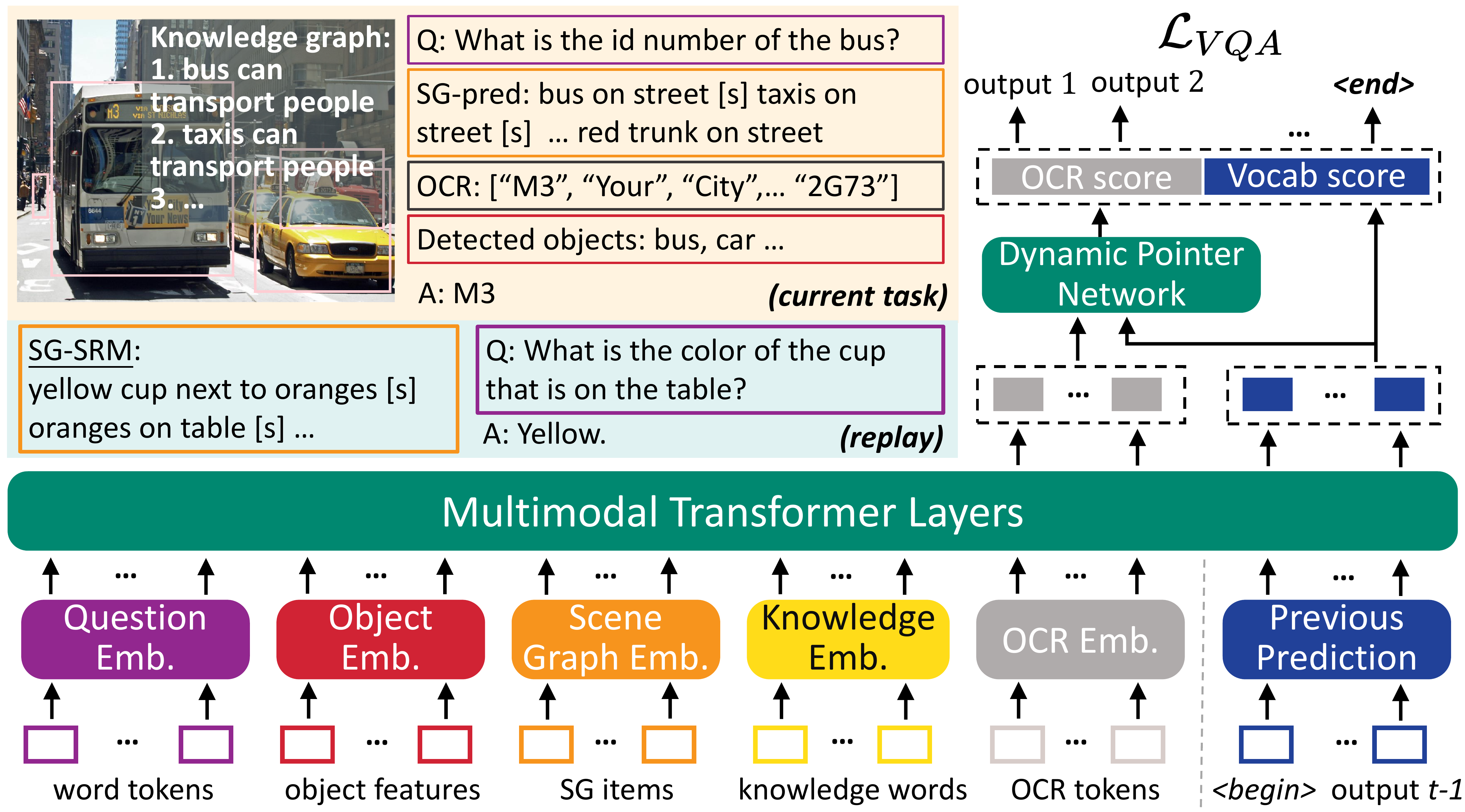} 
\caption{
Architecture of UniVQA. It extracts features of all inputs and projects them into a common space. Then, it applies multimodal transformer layers with dynamic pointer network to auto-regressively predict the answer, where each word could be an OCR token or a word in the vocabulary.
}
\label{fig:univqa}
\end{figure}

In CLVQA, a VQA model should be able to continuously learn to tackle different types of questions. As a result, it may encounter inputs of different modalities at different stages. \textit{E.g.}, a VQA model might need to read the text and copy the OCR token for answering a question after it is taught how to verify an object’s attribute. Therefore, we propose a general VQA model based on Multi-modal Transformer~\cite{hu2020iterative}. Details are shown in~\cref{fig:univqa}.
The UniVQA model $S$ supports various type of inputs, including images, questions, text-form scene graphs, knowledge and OCR. Using this information from the inputs, it learns to predict the answer.
Following M4C~\cite{hu2020iterative}, we develop a multimodal transformer to serve as the VQA backbone, which generalizes to various input modalities.
Concretely, given the input information, we extract feature representations of different modalities and then 
 project each type of feature into a common $d$-dimensional semantic space. 
 A multi-layer transformer is then used to model and enrich the context of inter- and intra-modalities given the projected features. Finally, the model decodes the answer words iteratively with a dynamic pointer network. Note that an answer word could either yield from the answer vocabulary or be copied from some existing tokens(\textit{e.g.}, OCR tokens here).

\head{Feature extraction.} We extract features from general fields (e.g., question word features and object features) and task-specific fields (e.g., OCR token features from an external OCR system when tackling VQA needs reading texts in an image). 

For language-based feature extraction, we use a pre-trained BERT model~\cite{vaswani2017bert_attention} for extraction\footnote{We use the first 3 layers of BERT-BASE since it is sufficient and saves computation}, 
and fine-tune the BERT parameters via the question answering loss. We apply this extraction to question, scene graph, and knowledge.

For object feature extraction, we obtain a set of visual objects through the Faster R-CNN model~\cite{ren2015faster} and combine their appearance feature and location feature. 
The appearance feature and location feature of the detected object are linearly transformed into a $d$-dimensional space, followed by a layer normalization~\cite{ba2016layer}.

For OCR tokens, we follow~\cite{hu2020iterative} to obtain the feature by summing up the FastText vector~\cite{bojanowski2017enriching}, Faster R-CNN based RoI-Pooling feature~\cite{ren2015faster}, PyramidalHistogram of Characters (PHOC) vector and location feature, followed by a layer normalization~\cite{ba2016layer}.

\head{Multi-modal fusion and answer prediction.} 
After embedding and projecting features for all modalities, we apply stacked layers of transformers over the list of all features. 
Through the multi-head self-attention mechanism, each modality is allowed to freely attend to all others. 
The output from the multi-modal transformer layers is a list of $d$-dimensional feature vectors, which can be viewed as the enriched embedding in a multi-modal context.

Then we follow the design in~\cite{hu2020iterative} to decode answer words in an auto-regressive manner, where each decoded word could be either an OCR token copied by the dynamic pointer network, or a word from the answer vocabulary. 
At each decoding step, we feed in an embedding of the previously predicted word, and predict the next answer word based on the score of OCR tokens output from the dynamic pointer network and the score of words in the answer vocabulary. We augment the answer vocabulary with two special tokens, \texttt{<begin>} and \texttt{<end>}. \texttt{<begin>} is used as the input to the first decoding step, and we stop the decoding process after \texttt{<end>} is predicted.

We mask the attention weights in the self-attention layers of the multi-modal transformer to ensure causality in answer decoding. In this way, features of all input modalities cannot attend to any decoding steps, and all decoding steps can only attend to previous word features in addition to features of all input modalities.

\head{Data input scheme.}
In CLVQA, for the current task $T_i$, we extract object features, question words features, and some task-specific token features (\textit{e.g.}, OCR token features for \textit{scene text recognition} and knowledge feature for \textit{knowledge reasoning}). 
However, the replayed samples generated by $S_{i-1}$ do not contain images or visual features. 
To reduce the gap of inputs, we extract the scene graph of the image of the current task via an offline scene graph generation model~\cite{tang2020unbiased} , viewing it as plain text and extracting its language-based feature using BERT. We feed this feature of scene graph into the multi-modal transformer as well. 
For replayed samples, we extract the language feature for SG-SRM. 

\head{Training}. In task $T_i$ ($i > 1$), $U$ is trained on the mix of current samples and generated samples. In experiments, we generate $\gamma |D_i|$ samples from $S_{i-1}$ and all previous tasks have the same share. That is, when beginning training $U_i$, we generate $\frac{\gamma}{i-1}|D_i|$ samples for each of the previous $i-1$ tasks.
Formally, the loss function of the $i$-th VQA model is given as:
\begin{multline*}
    \mathcal{L}_{VQA}(\vv{\phi}) = \mathbb{E}_{\left\{ \ve{v}, \ve{q}, \ve{a} \right\}\sim D_i } 
    \left[
        \mathcal{L}\left( U\left(\ve{v},\ve{q}; \vv{\phi}\right), \ve{a} \right)
    \right] + \\
    \mathbb{E}_{\left\{ \ve{v}', \ve{q}', \ve{a}' \right\}\sim S_{i-1} } 
    \left[
        \mathcal{L}\left( U\left(\ve{v}',\ve{q}'; \vv{\phi}\right), \ve{a}' \right)
    \right],
\end{multline*}
where $\vv{\phi}$ is the model parameters of $U$ and $\mathcal{L}(\cdot)$ is the answer prediction loss.

\section{Experiments}
\label{sec:exp}
We proceed with the experimental assessment of the CLVQA task on CLOVE. In this section, we first discuss the baseline methods for CLVQA and implementation details of our method. Then we proceed with empirical assessment of these methods on the CLOVE benchmark.

\subsection{Baselines}
The following baselines are evaluated in our experiments:

\head{\# Finetune}: It finetunes the UniVQA model sequentially without any other modules, memory buffers or any other losses.

\head{\# EWC}~\cite{schwarz2018progress_onlineEWC}: We apply online EWC, a transformed version of EWC, which accumulates the importance of the stream of tasks.

\head{\# MAS}~\cite{aljundi2018mas}: A regularization based method estimating importance via the gradients of model outputs.

\head{\# LAMOL-m}~\cite{sun2019lamol}: LAMOL tackles CL in NLP by generating pseudo samples for experience replay using a language model. Here, we modify its pipeline to adapt to CLVQA task. Specifically, we apply MSE loss for object feature regression and use the same QA loss as in the original LAMOL.

\head{\# VQG}~\cite{krishna2019information}: It saves part of images and answers from previous tasks and use a visual question generation model to generate coherent questions for replay. 
A pseudo-replayed sample consists of an image and an answer from a previous task, and the generated question from VQG.

\head{\# Real data replay}: It finetunes model augmented with real samples saved into an episodic memory. We adopt two strategies to choose the real samples. (1) \textbf{Real-rnd}: randomly choose samples and update the memory with the latest samples; (2) \textbf{Real-Kmeans}~\cite{huang2021continual_diyi}: we add a learnable token for UniVQA and used its output feature from the multimodal transformer to represent the input sample of the current task.
We then use these features to conduct K-means and select the samples closest to each cluster’s centroid, following.
In the experiments, we set the memory buffer size to be equal to our SRM's model size plus the saved SG-prompts. See Supp. for details of all methods.

\subsection{Implementation details}
For SRM, we use the pre-trained distilGPT2 for model initialization at the beginning of a task sequence.
We use an AdamW~\cite{loshchilov2017adamw} optimizer with $lr = 6.25 \times  10^{-5}$, 
$\beta_1 = 0.9 $ and $\beta_2 = 0.99$. $\lambda$ is set to be 0.25. The pseudo samples and current task samples are mixed at a ratio of $1:4$. We train each task for 15 epochs. 
For UniVQA, we use $d=768$ as the dimension of the joint embedding space. We use 4 layers of multi-modal transformer with 12 attention heads and random initialize its parameters. Other hyper-parameters follow BERT-BASE~\cite{vaswani2017bert_attention}. We set the maximum decoding steps to be 12 in answer prediction. During training, we set the batch size to be 32 and train the model for a maximum of 24,000 iterations. We use the Adam optimizer, with $lr = 1.0 \times 10^{-4}$ and a staircase learning rate schedule, where we multiply the learning rate by 0.1 at 14,000 and at 19,000 iterations. 

Note that UniVQA and other baselines use the same training protocol. \textbf{\# EWC} and \textbf{\# MAS} are implemented upon the base model of UniVQA. For \textbf{\# LAMOL-m} and \textbf{\# VQG}, the replayed samples are mixed with the current task samples to jointly train at current task.

\begin{table*}[h]
\centering
\resizebox{\textwidth}{!}{%
\begin{tabular}{lccccccccccccccc}
\toprule
\multirow{2}{*}{Method} & \multicolumn{7}{c}{CLOVE-scene}                             & \multicolumn{1}{l}{} & \multicolumn{7}{c}{CLOVE-function}                          \\ \cline{2-8} \cline{10-16} 
                        & \textit{abcdef} & \textit{bdfcae} & \textit{beacfd} & \textit{beadcf} & \textit{bedfca} & \textit{ecdfab} & Avg.   &                      & \textit{oarlks} & \textit{roslak} & \textit{rklsao} & \textit{rsolak} & \textit{lkosra} & \textit{kaorls} & Avg.   \\ \cline{1-8} \cline{10-16} 
Finetune                & 27.53  & 27.98  & 28.39  & 27.71  & 24.49  & 25.42  & 26.92 &                      & 27.60  & 29.33  & 21.12  & 30.65  & 25.43  & 22.82  & 26.16 \\
EWC                    & 27.59  & 27.64  & 28.47  & 29.18  & 24.03  & 25.48  & 27.07 &                      & 29.26  & 30.87  & 21.87  & 28.69  & 23.58  & 23.27  & 26.26 \\
MAS                  & 27.41  & 27.15  & 28.19  & 27.34  & 25.40  & 26.78  & 27.05 &                      & 28.73  & 31.59  & 28.62  & 28.57  & 24.26  & 26.73  & 28.08 \\
VQG                   & 29.15  & 29.74  & 30.02  & 30.27  & 27.28  & 28.66  & 29.19 &                      & 32.78  & 33.16  & 29.55  & 33.82  & 30.17  & 28.67  & 31.36 \\
LAMOL-m                & 29.40  & 28.52  & 29.45  & 29.86  & 26.52  & 27.82  & 28.60 &                      & 28.42  & 29.04  & 24.16  & 32.17  & 26.94  & 26.92  & 27.94 \\
\textbf{SGP (Ours)} &
  \textbf{32.21} &
  \textbf{33.72} &
  \textbf{34.37} &
  \textbf{33.18} &
  \textbf{31.84} &
  \textbf{32.98} &
  \textbf{33.05} &
  \textbf{} &
  \textbf{45.97} &
  \textbf{41.80} &
  \textbf{39.05} &
  \textbf{42.95} &
  \textbf{38.65} &
  \textbf{43.62} &
  \textbf{42.01} \\ \cmidrule{1-8} \cmidrule{10-16} 
{\color[HTML]{9B9B9B} Real-rnd } & {\color[HTML]{9B9B9B} 36.60 } & {\color[HTML]{9B9B9B} 37.69 } & {\color[HTML]{9B9B9B} 35.50 } & {\color[HTML]{9B9B9B} 36.51 } & {\color[HTML]{9B9B9B} 35.86 } & {\color[HTML]{9B9B9B} 36.84 } & {\color[HTML]{9B9B9B} 36.50 } & {\color[HTML]{9B9B9B}  } & {\color[HTML]{9B9B9B} 44.83 } & {\color[HTML]{9B9B9B} 42.62 } & {\color[HTML]{9B9B9B} 39.28 } & {\color[HTML]{9B9B9B} 43.37 } & {\color[HTML]{9B9B9B} 40.85 } & {\color[HTML]{9B9B9B} 40.08 } & {\color[HTML]{9B9B9B} 41.84 }\\
{\color[HTML]{9B9B9B} Real-kmeans } & {\color[HTML]{9B9B9B} 36.91 } & {\color[HTML]{9B9B9B} 38.15 } & {\color[HTML]{9B9B9B} 37.01 } & {\color[HTML]{9B9B9B} 38.30 } & {\color[HTML]{9B9B9B} 37.93 } & {\color[HTML]{9B9B9B} 34.86 } & {\color[HTML]{9B9B9B} 37.19 } & {\color[HTML]{9B9B9B}  } & {\color[HTML]{9B9B9B} 40.28 } & {\color[HTML]{9B9B9B} 41.19 } & {\color[HTML]{9B9B9B} 38.49 } & {\color[HTML]{9B9B9B} 42.21 } & {\color[HTML]{9B9B9B} 38.39 } & {\color[HTML]{9B9B9B} 36.29 } & {\color[HTML]{9B9B9B} 39.48 }\\
\cmidrule{1-8} \cmidrule{10-16} 
Offline                 & \multicolumn{7}{c}{48.45}                                   & \multicolumn{1}{r}{} & \multicolumn{7}{c}{57.53}                                   \\ \bottomrule
\end{tabular}%
}
\caption{Summary of average accuracy(\%) for different methods under six task orders in CLOVE-scene and CLOVE-function respectively. We use models at last iteration of last task for testing. \textbf{Offline} training as an upper bound is shown at the bottom. Our method outperforms all other real-data-free baselines and achieves comparable performance with real data replay. Rows with grey fonts use the real data while replaying.}

\label{tab:main_results}
\end{table*}

\subsection{Results and Analyses}
\head{Experiments on different task orders.} 
For both CLOVE-scene and CLOVE-function settings, we randomly sample 6 task orders from all possible permutations for evaluation. We use the following abbreviation scheme to simplify the task order notation: \textit{oarlks} denotes \textit{object recognition} $\rightarrow$ \textit{attribute recognition }$\rightarrow$ \textit{relation reasoning} $\rightarrow$ \textit{logic reasoning }$\rightarrow$ \textit{knowledge reasoning }$\rightarrow$ \textit{scene text recognition} and \textit{abcdef} denotes \textit{ShopAndDining} $\rightarrow$ \textit{Workplace} $\rightarrow$ \textit{HomeorHotel} $\rightarrow$ \textit{Transportation} $\rightarrow$ \textit{SportAndLeisure} $\rightarrow$ \textit{Outdoors}. Besides, for both settings, we set $\gamma=1.5$ for SGP and report the average accuracy.

From the results in~\cref{tab:main_results}, we can find that our method outperforms baselines and SOTA methods on other CL tasks without saving real data by a large margin. Compared to LAMOL-m, our SRM might replay better visual representation than that of LAMOL-m, where object feature regression is applied.
Besides, although VQG additionally saves real images, answers and the corresponding task labels as input to generate questions, our SRM still obviously outperforms the VQG. This is probably because by using the scene graph, we can better capture the symbolic relations between visual content and QA pairs. The model can better learn to ask questions by using scene graphs than images. 

Moreover, compared to real data replay methods, although they save the real data, SGP still achieves comparable performance under the functional setting and is even slightly better on some task orders. Note that on other CL benchmarks, the pseudo-replay \textit{sota} methods usually still has a non-negligible gap with the real-replay methods~\cite{van2018generative,sun2019lamol}. Thus, we believe that symbolic replay could be a very promising direction.
Under the CLOVE-scene, our method underperforms real data replay methods. The reason could be that the disjoint image split in CLOVE-scene increases the difficulty in combating the forgetting in visual modality. Thus, the replayed scene graphs generated by SRM are less informative than real images in retaining visual knowledge.

We also find that the performances of all previous methods without using real data are not satisfactory.
In some task orders, regularization-based methods (\textit{i.e.}, EWC \& MAS) perform even worse than Finetune. It indicates that simply measuring the importance of multi-modal network's parameters might fail to decouple the forgetting in complex multiple modalities and lead to a sub-optimal regularization. 
LAMOL-m and VQG outperform Finetune in both settings, indicating that the pseudo-replay based methods work for CLVQA, while their improvements are relatively limited.

For the real data replay baselines, we can find that Real-kmeans works better under the CLOVE-scene, while in CLOVE-function, the naive real-rnd works slightly better. We conjecture that Kmeans relies on discriminative sample representations. Image features used in previous works are usually well discriminative, but complex multi-modal VQA features may not, especially for CLOVE-function, which involves more diverse types of multi-modal reasoning. This shows that for sample selection methods, how to represent a VQA sample is still an open problem.

\head{Importance of scene graph replay.} We remove the next token prediction task when training SRM, and only generate QA pairs during replay. Comparing \#1 and \#2 in~\cref{tab:abl_gtsg}, we notice an obvious drop in final average accuracy when scene graphs aren't replayed.
It shows that our replayed scene graph can effectively retain visual contexts from previous tasks. In supplementary, we show some data examples generated by our SRM.

\head{Using GT scene graph for prompting.} As mentioned before, for symbolic replay in SRM, we randomly sample a few scene graph items as a prompt to complete the scene graph and generates corresponding QA pairs. 
Here, we would like to see the effect of the random prompt sampling on CLVQA. Thus, we propose a variant of SRM, which saves GT scene graphs used to generate QA pairs after training each task and samples the GT scene graph prompts for symbolic replay.
Comparing \#2 and \#3 in~\cref{tab:abl_gtsg}, SGM with GT prompts can boost the average accuracy by 3.01\% and 2.8\% in CLOVE-scene and CLOVE-function settings respectively. It indicates that improving the sampling strategy could be one direction to obtain better CL performance.

\begin{table}[h]
\centering
\resizebox{0.47\textwidth}{!}{%
\begin{tabular}{ccccc}
\toprule
No. & Prompt type  & Replay elements & CLOVE-Scene & CLOVE-Function \\ \midrule
\#1   &   Random         & Q + A          & 29.52 & 40.24    \\
\#2   &   Random         & SG + Q + A     & 32.08 & 44.21    \\
\#3   &   GT          & SG + Q + A     & 35.09   &  47.01   \\
 \bottomrule
\end{tabular}
}
\caption{Comparison of different types of SG-prompt and replay elements (reported under $\gamma=0.9$ on average accuracy (\%)).} 
\label{tab:abl_gtsg}
\end{table}

\begin{figure}[h!]
\centering
\includegraphics[width=\columnwidth]{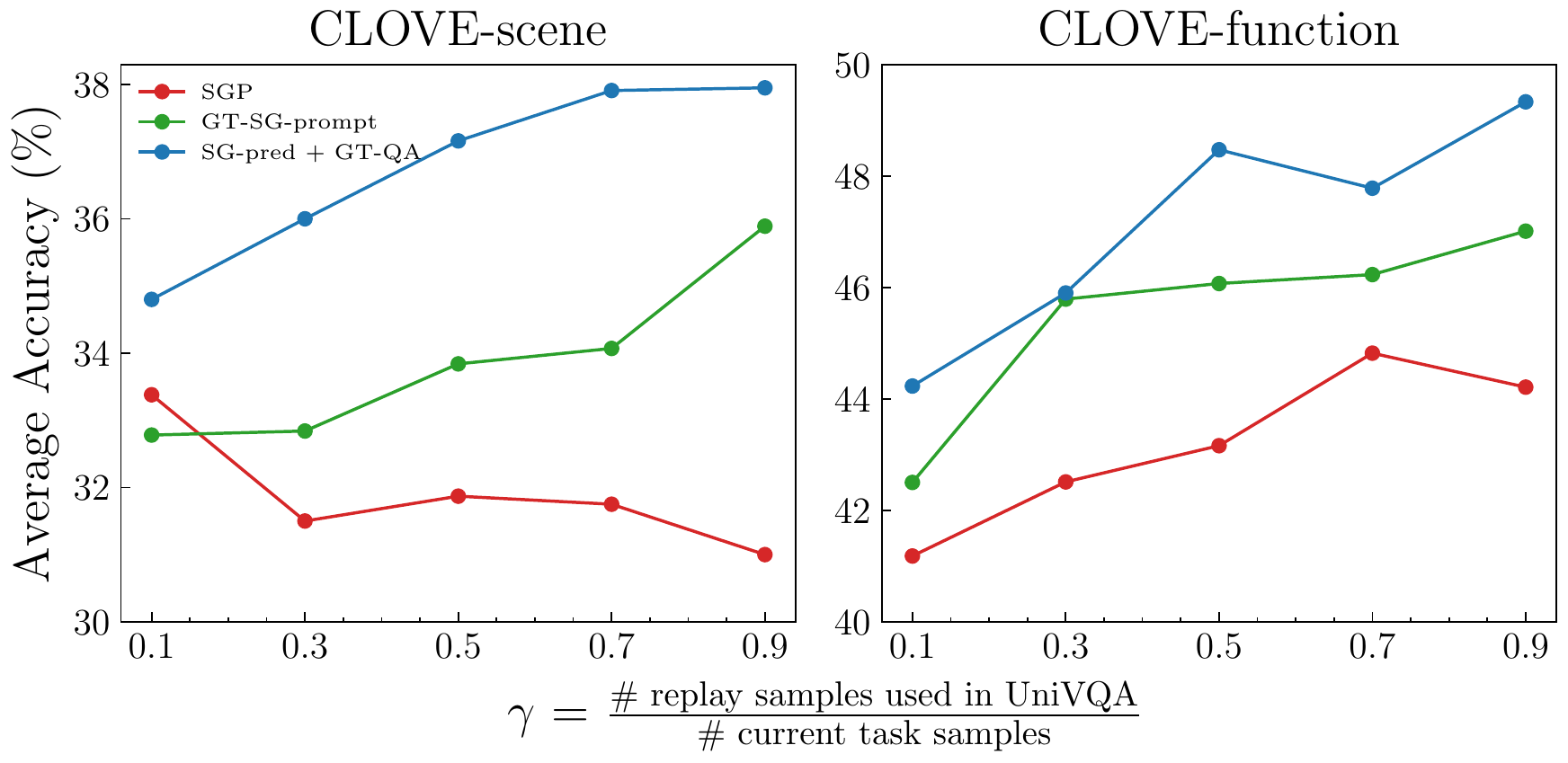} %
\caption{ Comparing different replay schemes of SGP. ``SGP'' denotes our SGP using randomly sampled SG-prompt to generate scene graph, question and answer. ``GT-SG-prompt'' denotes our SGP using ground-truth SG-prompt to generate scene graph, question and answer. ``SG-pred + GT-QA'' denotes saving the predicted scene graph of an image by an offline scene graph predictor, and ground truth question and answer. 
}
\label{fig:supp_gamma}
\end{figure}

\head{Potential of SGP when generating more precise QA pairs.} To explore how well an ideal SGP can combat forgetting in CLOVE-scene and CLOVE-function, we compare our SGP with (1) SGP prompt by GT SG-prompt, which means that we save the ground truth scene graph relationship for question generation from the training set of each task, and use the saved SG-prompt as prefix for replay. (2) Replay data with saved predicted scene graph from an offline scene graph predictor, and ground truth QA pair. This can be viewed as a pseudo upper bound of SGP as it replays the real QA pair and corresponding scene graph. 
We should the results of $\gamma = [0.1, 0.3, 0.5, 0.7, 0.9]$ under \textit{abcdef} in CLOVE-scene and \textit{oarlks} in CLOVE-function.
Results are shown in~\cref{fig:supp_gamma}. We can observe that for both the CLOVE-scene and CLOVE setting, under different numbers of replayed samples ($\gamma$), (1). Using ground-truth scene graphs outperforms using randomly sampled SG-prompt, indicating that sampling or sourcing better SG-prompt could be a potential solution. (2). Saving scene graphs, questions and answers outperforms all others. This finding indicates that replaying an oracle \textit{scene-graph-quesiton-answer} triplet could further enhance the performance, and this can be a promising direction for real-data-free replay-based method on CLVQA.

\begin{figure}[ht]
\centering
\includegraphics[width=\columnwidth]{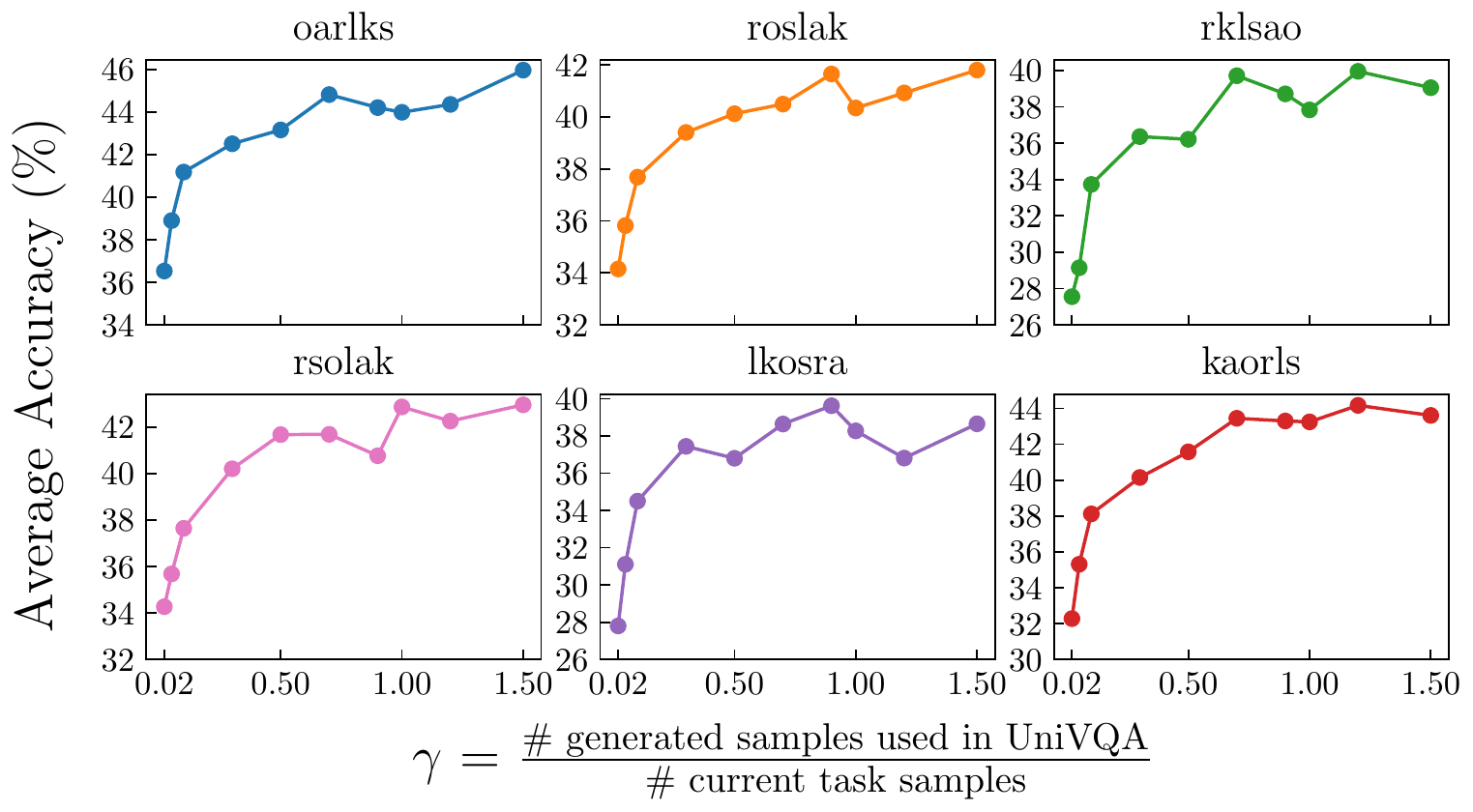}
\caption{Results of using different number of generated samples in training UniVQA on CLOVE-function.
}
\label{fig:gamma}
\end{figure}

\head{Using different number of generated samples in training UniVQA.} We set different $\gamma$ in this experiment to inspect the impact of adopting different amounts of generated samples in training UniVQA.
Fig.~\ref{fig:gamma} illustrates that the performance is relatively low at small $\gamma$, indicating that a small number of replayed data might not be sufficient against forgetting. Also, the performance reaches stable when $\gamma > 0.7$. The reason could be that when the sample reaches a certain number, generating more data will not increase the diversity of the samples, but lead to imbalanced distribution of the replay and current samples and further jeopardize the performance.

\begin{figure}[ht]
\centering
\includegraphics[width=0.99\columnwidth]{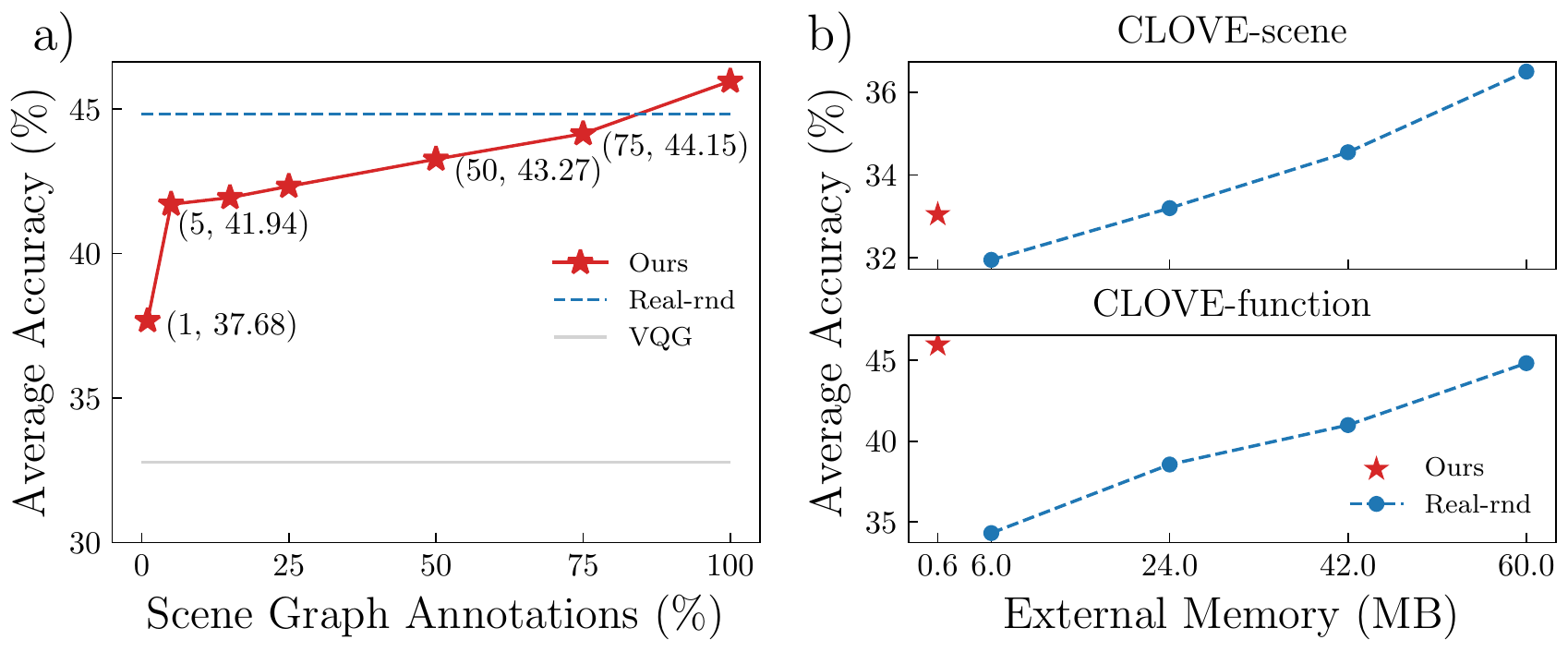}
\caption{a) Results of using different proportional annotations in training SRM under CLOVE-function-\textit{oarlks}.
b) Compare SGP with Real-rnd with different memory size.
}
\label{fig:abl}
\end{figure}

\head{Using less scene graph annotations in training SRM.} SRM requires scene graph annotation, which may need extra annotation costs. However, here, we would like to illustrate that SRM is not data-hungry. It can achieve good results with just a small amount of data. Specifically, we reduce the annotations used in SRM training by randomly sampling $r|D_i|$ samples for each task. We showcase the result in~\cref{fig:abl} (a). We can see that only using 50\% data can lead to an average accuracy of 43.27\%, which is still close to the Real-rnd method. Also, the SRM with only 1\% training data (i.e., only about 200 VQA samples) still outperforms VQG, the best model without saving real data, by 4.9\%.


\head{Comparing external memory size.} We compare our method with the real data replay method with different memory sizes for saving data. For our method, our SGP only needs to save the sampled prompts, so we calculate the memory size of saving the largest number of samples in our experiments (\textit{i.e.}, $\gamma=1.5$). For real data replay, we adopt memory of different storage sizes. As is shown in~\cref{fig:abl} (b), our method is memory-efficient in terms of saving external data: by saving only 612KB SG-prompts, it achieves comparable performance with Real-rnd saving 24MB (40.2$\times$) real data under CLOVE-scene setting, and Real-rnd saving 60MB (100.4$\times$) real data under CLOVE-function setting.

\section{Conclusion}
This paper proposes CLOVE benchmark investigating the CL of VQA under scene- and function-incremental settings. We also propose a framework, Scene Graph as Prompt for symbolic replay, which retains past knowledge by using replayed scene graphs and correlated QA pairs. Extensive experiments show the superiority of our method over other real-data-free CL methods. Besides, we find that previous CL methods face several unique challenges in CLOVE: 1) How regularization methods decouple multi-modal information from network's parameters. 2) How real data replay methods obtain a discriminative representation of a VQA sample suitable for sample selection. 3) How pseudo replay methods generate plausible images or alternatives of images. Finally, we hope CLOVE can provide an enabling resource to build VQA systems with powerful CL ability.

\begin{table*}[h]
\centering
\resizebox{\textwidth}{!}{%
\begin{tabular}{cccccccccccccccccccccccc}
\hline
 & \multicolumn{3}{c}{\textit{abcdef}} &  & \multicolumn{3}{c}{\textit{bdfcae}} &  & \multicolumn{3}{c}{\textit{beacfd}} &  & \multicolumn{3}{c}{\textit{beadcf}} &  & \multicolumn{3}{c}{\textit{bedfca}} &  & \multicolumn{3}{c}{\textit{ecdfab}} \\ \cline{2-4} \cline{6-8} \cline{10-12} \cline{14-16} \cline{18-20} \cline{22-24} 
\multirow{-2}{*}{Method} & A ($\uparrow$) & F ($\downarrow$) & B ($\uparrow$) &  & A ($\uparrow$) & F ($\downarrow$) & B ($\uparrow$) &  & A ($\uparrow$) & F ($\downarrow$) & B ($\uparrow$) &  & A ($\uparrow$) & F ($\downarrow$) & B ($\uparrow$) &  & A ($\uparrow$) & F ($\downarrow$) & B ($\uparrow$) &  & A ($\uparrow$) & F ($\downarrow$) & B ($\uparrow$) \\ \cline{1-4} \cline{6-8} \cline{10-12} \cline{14-16} \cline{18-20} \cline{22-24} 
Finetune & 27.53 & 32.92 & -24.60 &  & 27.98 & 35.14 & -23.44 &  & 28.39 & 33.79 & -23.21 &  & 27.71 & 31.62 & -23.92 &  & 24.49 & 36.21 & -28.19 &  & 25.42 & 35.50 & -25.95 \\
EWC & 27.59 & 31.12 & -23.46 &  & 27.64 & 34.57 & -24.11 &  & 28.47 & 34.31 & -23.07 &  & 29.18 & 30.14 & -22.47 &  & 24.03 & 36.59 & -28.85 &  & 25.48 & 35.21 & -25.72 \\
MAS & 27.41 & 31.73 & -23.69 &  & 27.15 & 36.22 & -24.60 &  & 28.19 & 34.98 & -23.95 &  & 27.34 & 31.79 & -24.36 &  & 25.40 & 34.19 & -26.71 &  & 26.78 & 33.35 & -24.50 \\
VQG & 29.15 & 29.57 & -21.73 &  & 29.74 & 32.46 & -22.02 &  & 30.02 & 32.71 & -20.32 &  & 30.27 & 28.82 & -19.88 &  & 27.28 & 31.92 & -24.59 &  & 28.66 & 31.48 & -22.72 \\
LAMOL-m & 29.40 & 29.32 & -21.47 &  & 28.52 & 33.58 & -23.26 &  & 29.45 & 33.12 & -20.59 &  & 29.86 & 29.57 & -20.27 &  & 26.52 & 32.86 & -25.80 &  & 27.82 & 32.16 & -23.91 \\
\textbf{SGP (Ours)} & \textbf{32.21} & \textbf{26.51} & \textbf{-18.94} & \textbf{} & \textbf{33.72} & \textbf{27.56} & \textbf{-16.60} & \textbf{} & \textbf{34.37} & \textbf{27.40} & \textbf{-15.75} & \textbf{} & \textbf{33.18} & \textbf{26.27} & \textbf{-17.11} & \textbf{} & \textbf{31.84} & \textbf{27.27} & \textbf{-18.97} & \textbf{} & \textbf{32.98} & \textbf{27.41} & \textbf{-17.31} \\ \cline{1-4} \cline{6-8} \cline{10-12} \cline{14-16} \cline{18-20} \cline{22-24} 
 
{\color[HTML]{9B9B9B} Real-rnd } & {\color[HTML]{9B9B9B} 36.60 } & {\color[HTML]{9B9B9B} 21.21 } & {\color[HTML]{9B9B9B} -13.43 } & {\color[HTML]{9B9B9B}  } & {\color[HTML]{9B9B9B} 37.69 } & {\color[HTML]{9B9B9B} 23.60 } & {\color[HTML]{9B9B9B} -11.83 } & {\color[HTML]{9B9B9B}  } & {\color[HTML]{9B9B9B} 35.50 } & {\color[HTML]{9B9B9B} 27.16 } & {\color[HTML]{9B9B9B} -14.57 } & {\color[HTML]{9B9B9B}  } & {\color[HTML]{9B9B9B} 36.51 } & {\color[HTML]{9B9B9B} 21.03 } & {\color[HTML]{9B9B9B} -13.25 } & {\color[HTML]{9B9B9B}  } & {\color[HTML]{9B9B9B} 35.86 } & {\color[HTML]{9B9B9B} 22.09 } & {\color[HTML]{9B9B9B} -14.08 } & {\color[HTML]{9B9B9B}  } & {\color[HTML]{9B9B9B} 36.84 } & {\color[HTML]{9B9B9B} 21.34 } & {\color[HTML]{9B9B9B} -12.27 } \\

{\color[HTML]{9B9B9B} Real-kmeans } & {\color[HTML]{9B9B9B} 36.91 } & {\color[HTML]{9B9B9B} 21.73 } & {\color[HTML]{9B9B9B} -13.32 } & {\color[HTML]{9B9B9B}  } & {\color[HTML]{9B9B9B} 38.15 } & {\color[HTML]{9B9B9B} 22.21 } & {\color[HTML]{9B9B9B} -11.14 } & {\color[HTML]{9B9B9B}  } & {\color[HTML]{9B9B9B} 37.01 } & {\color[HTML]{9B9B9B} 24.26 } & {\color[HTML]{9B9B9B} -12.94 } & {\color[HTML]{9B9B9B}  } & {\color[HTML]{9B9B9B} 38.30 } & {\color[HTML]{9B9B9B} 19.61 } & {\color[HTML]{9B9B9B} -11.41 } & {\color[HTML]{9B9B9B}  } & {\color[HTML]{9B9B9B} 37.93 } & {\color[HTML]{9B9B9B} 20.23 } & {\color[HTML]{9B9B9B} -12.01 } & {\color[HTML]{9B9B9B}  } & {\color[HTML]{9B9B9B} 34.86 } & {\color[HTML]{9B9B9B} 24.71 } & {\color[HTML]{9B9B9B} -14.73 }
\\ \hline
\end{tabular}%
}
\caption{Summary of average accuracy (A), forgetting (F) and backward transfer (B) for different under six task orders in CLOVE-scene. $\uparrow$ means higher is better; $\downarrow$ means lower is better. Rows with grey fonts use the real data for replay.}
\label{tab:suppl_results_scene}
\end{table*}

\begin{table*}[h]
\centering
\resizebox{\textwidth}{!}{%
\begin{tabular}{cccccccccccccccccccccccc}
\hline
 & \multicolumn{3}{c}{\textit{oarlks}} &  & \multicolumn{3}{c}{\textit{roslak}} &  & \multicolumn{3}{c}{\textit{rklsao}} &  & \multicolumn{3}{c}{\textit{rsolak}} &  & \multicolumn{3}{c}{\textit{lkosra}} &  & \multicolumn{3}{c}{\textit{kaorls}} \\ \cline{2-4} \cline{6-8} \cline{10-12} \cline{14-16} \cline{18-20} \cline{22-24} 
\multirow{-2}{*}{Method} & A ($\uparrow$) & F ($\downarrow$) & B ($\uparrow$) &  & A ($\uparrow$) & F ($\downarrow$) & B ($\uparrow$) &  & A ($\uparrow$) & F ($\downarrow$) & B ($\uparrow$) &  & A ($\uparrow$) & F ($\downarrow$) & B ($\uparrow$) &  & A ($\uparrow$) & F ($\downarrow$) & B ($\uparrow$) &  & A ($\uparrow$) & F ($\downarrow$) & B ($\uparrow$) \\ \cline{1-4} \cline{6-8} \cline{10-12} \cline{14-16} \cline{18-20} \cline{22-24} 
Finetune & 27.60 & 49.66 & -34.55 &  & 29.33 & 50.12 & -31.19 &  & 21.12 & 58.08 & -41.37 &  & 30.65 & 48.83 & -30.70 &  & 25.43 & 48.44 & -36.97 &  & 22.82 & 55.53 & -40.65 \\
EWC & 29.26 & 47.94 & -32.87 &  & 30.87 & 48.18 & -29.75 &  & 21.87 & 56.92 & -40.46 &  & 28.69 & 50.44 & -32.89 &  & 23.58 & 50.52 & -37.58 &  & 23.27 & 53.54 & -38.87 \\
MAS & 28.73 & 48.71 & -33.95 &  & 31.59 & 48.06 & -29.02 &  & 28.62 & 48.56 & -32.39 &  & 28.57 & 51.2 & -33.74 &  & 24.26 & 49.93 & -36.85 &  & 26.73 & 49.34 & -35.06 \\
VQG & 32.78 & 43.86 & -30.12 &  & 33.16 & 45.97 & -27.32 &  & 29.55 & 47.93 & -31.70 &  & 33.82 & 45.98 & -26.73 &  & 30.17 & 44.12 & -32.26 &  & 28.67 & 46.81 & -33.57 \\
LAMOL-m & 28.42 & 48.52 & -34.28 &  & 29.04 & 50.86 & -31.78 &  & 24.16 & 53.84 & -38.52 &  & 32.17 & 44.02 & -27.12 &  & 26.94 & 47.80 & -35.89 &  & 26.92 & 48.88 & -35.18 \\
\textbf{SGP (Ours)} & \textbf{45.97} & \textbf{27.39} & \textbf{-12.03} & \textbf{} & \textbf{41.80} & \textbf{36.22} & \textbf{-17.85} & \textbf{} & \textbf{39.05} & \textbf{36.50} & \textbf{-20.84} & \textbf{} & \textbf{42.95} & \textbf{33.78} & \textbf{-16.07} & \textbf{} & \textbf{38.65} & \textbf{32.38} & \textbf{-20.37} & \textbf{} & \textbf{43.62} & \textbf{31.20} & \textbf{-15.83} \\ \cline{1-4} \cline{6-8} \cline{10-12} \cline{14-16} \cline{18-20} \cline{22-24} 
 
{\color[HTML]{9B9B9B} Real-rnd } & {\color[HTML]{9B9B9B} 44.83 } & {\color[HTML]{9B9B9B} 29.55 } & {\color[HTML]{9B9B9B} -13.85 } & {\color[HTML]{9B9B9B}  } & {\color[HTML]{9B9B9B} 42.62 } & {\color[HTML]{9B9B9B} 34.53 } & {\color[HTML]{9B9B9B} -15.34 } & {\color[HTML]{9B9B9B}  } & {\color[HTML]{9B9B9B} 39.28 } & {\color[HTML]{9B9B9B} 36.81 } & {\color[HTML]{9B9B9B} -19.87 } & {\color[HTML]{9B9B9B}  } & {\color[HTML]{9B9B9B} 43.37 } & {\color[HTML]{9B9B9B} 33.57 } & {\color[HTML]{9B9B9B} -14.94 } & {\color[HTML]{9B9B9B}  } & {\color[HTML]{9B9B9B} 40.85 } & {\color[HTML]{9B9B9B} 29.91 } & {\color[HTML]{9B9B9B} -17.68 } & {\color[HTML]{9B9B9B}  } & {\color[HTML]{9B9B9B} 40.08 } & {\color[HTML]{9B9B9B} 35.00 } & {\color[HTML]{9B9B9B} -19.88 } \\

{\color[HTML]{9B9B9B} Real-kmeans } & {\color[HTML]{9B9B9B} 40.28 } & {\color[HTML]{9B9B9B} 30.80 } & {\color[HTML]{9B9B9B} -15.03 } & {\color[HTML]{9B9B9B}  } & {\color[HTML]{9B9B9B} 41.19 } & {\color[HTML]{9B9B9B} 36.35 } & {\color[HTML]{9B9B9B} -12.01 } & {\color[HTML]{9B9B9B}  } & {\color[HTML]{9B9B9B} 38.49 } & {\color[HTML]{9B9B9B} 37.71 } & {\color[HTML]{9B9B9B} -15.35 } & {\color[HTML]{9B9B9B}  } & {\color[HTML]{9B9B9B} 42.21 } & {\color[HTML]{9B9B9B} 35.34 } & {\color[HTML]{9B9B9B} -12.12 } & {\color[HTML]{9B9B9B}  } & {\color[HTML]{9B9B9B} 38.39 } & {\color[HTML]{9B9B9B} 32.61 } & {\color[HTML]{9B9B9B} -14.70 } & {\color[HTML]{9B9B9B}  } & {\color[HTML]{9B9B9B} 36.29 } & {\color[HTML]{9B9B9B} 34.21 } & {\color[HTML]{9B9B9B} -17.92 }
\\ \hline
\end{tabular}%
}
\caption{Summary of average accuracy (A), forgetting (F) and backward transfer (B) for different under six task orders in CLOVE-function. $\uparrow$ means higher is better; $\downarrow$ means lower is better. Rows with grey fonts use the real data for replay.}
\label{tab:suppl_results_func}
\end{table*}

\begin{figure*}[h!]
\centering
\includegraphics[width=\textwidth]{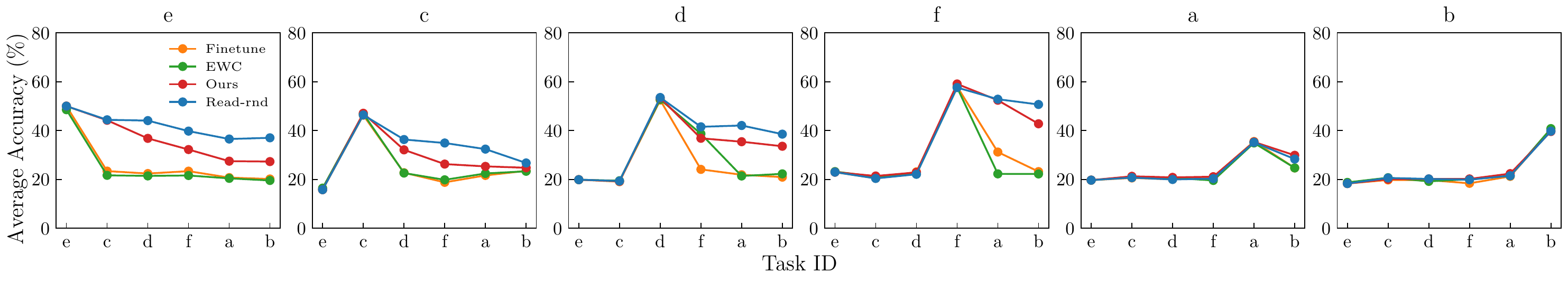} %
\caption{
The accuracy of different methods on each task of CLOVE-scene-\textit{ecdfab} as the training proceeds. 
}
\label{fig:supp_scene_ecdfab}
\end{figure*}

\begin{figure*}[h!]
\centering
\includegraphics[width=\textwidth]{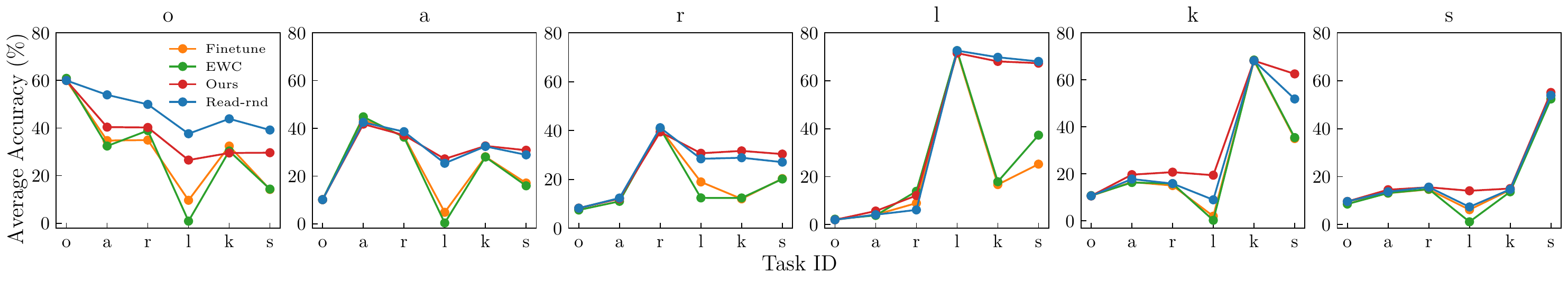} %
\caption{
The accuracy of different methods on each task of CLOVE-scene-\textit{oarlks} as the training proceeds.
}
\label{fig:supp_func_oarlks}
\end{figure*}

\appendix

\section*{Apendix}

\section{More Details of the CLOVE Benchmark}
\subsection{Answer Distribution}
\begin{figure*}[ht]
\centering
	
	\includegraphics[width=0.45\textwidth]{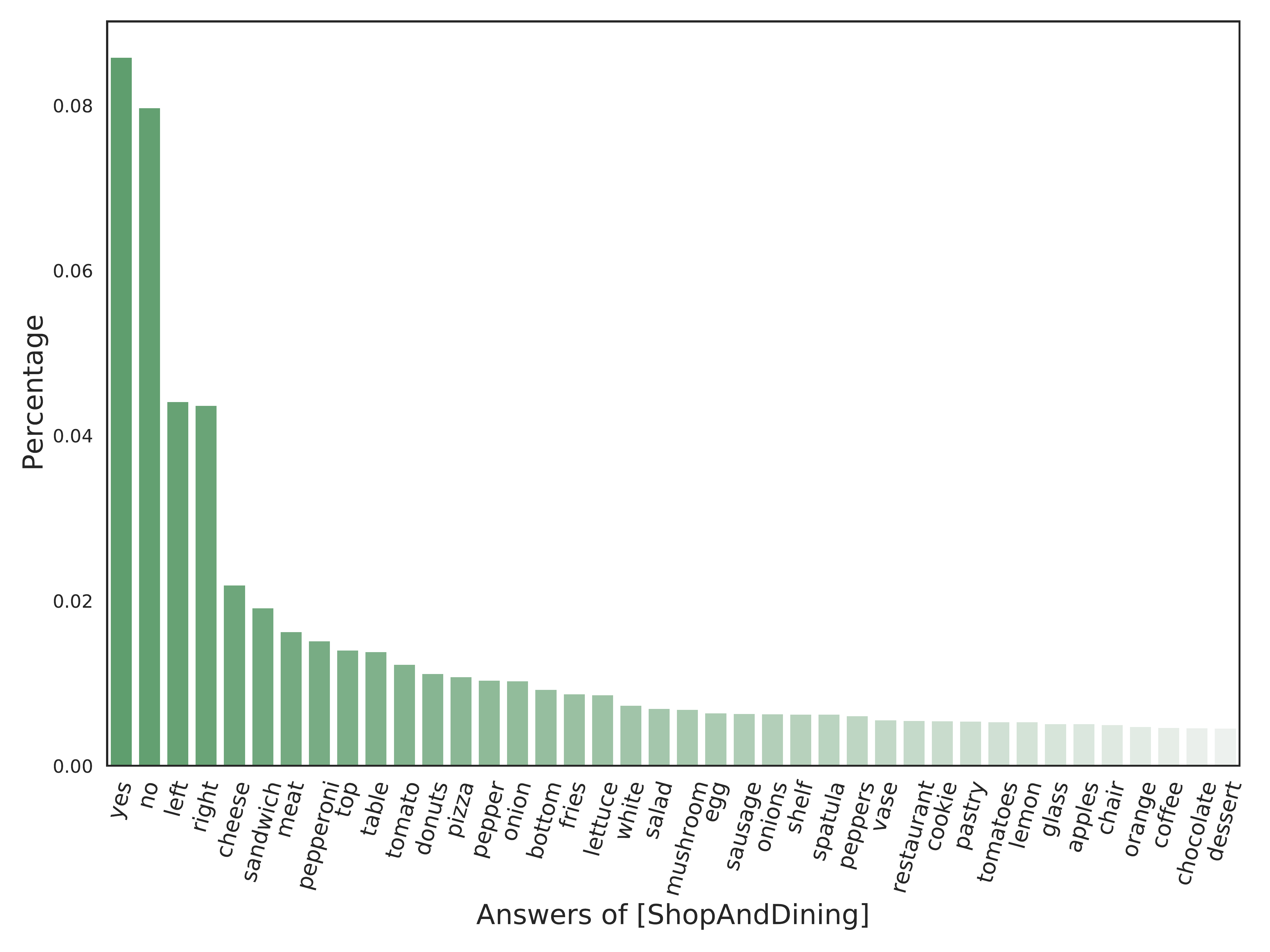}
	\includegraphics[width=0.45\textwidth]{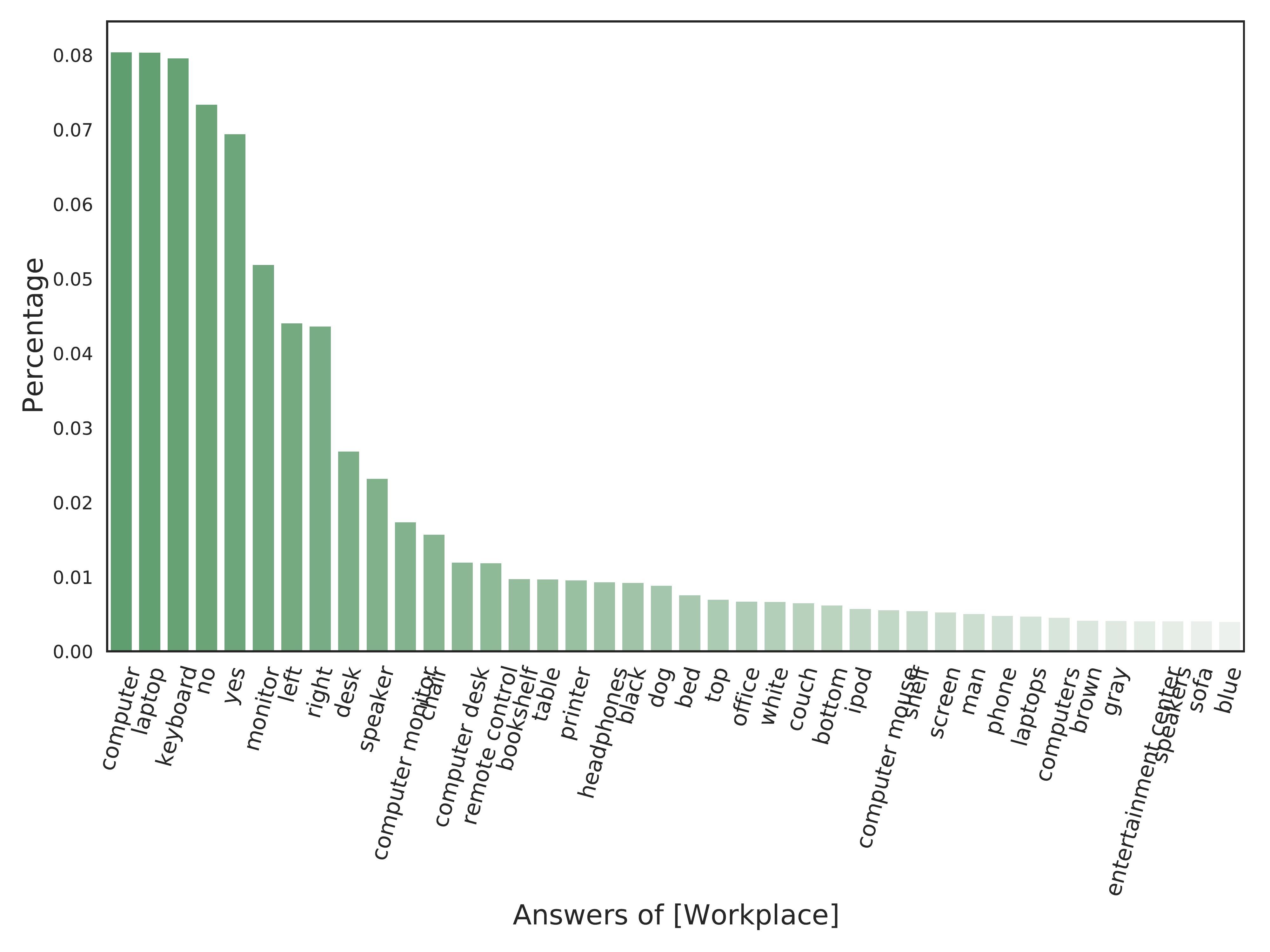} \\
	\includegraphics[width=0.45\textwidth]{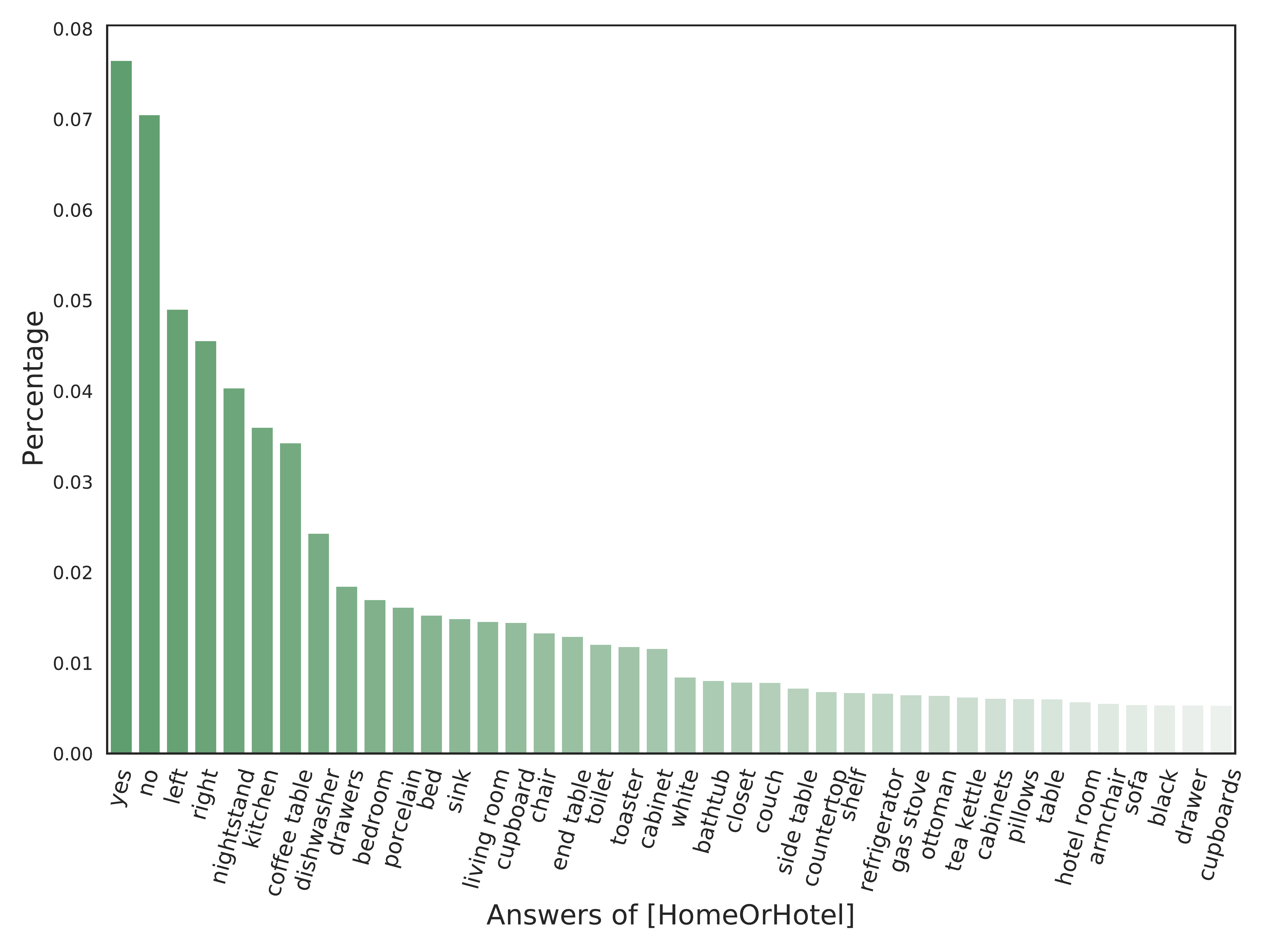}
	\includegraphics[width=0.45\textwidth]{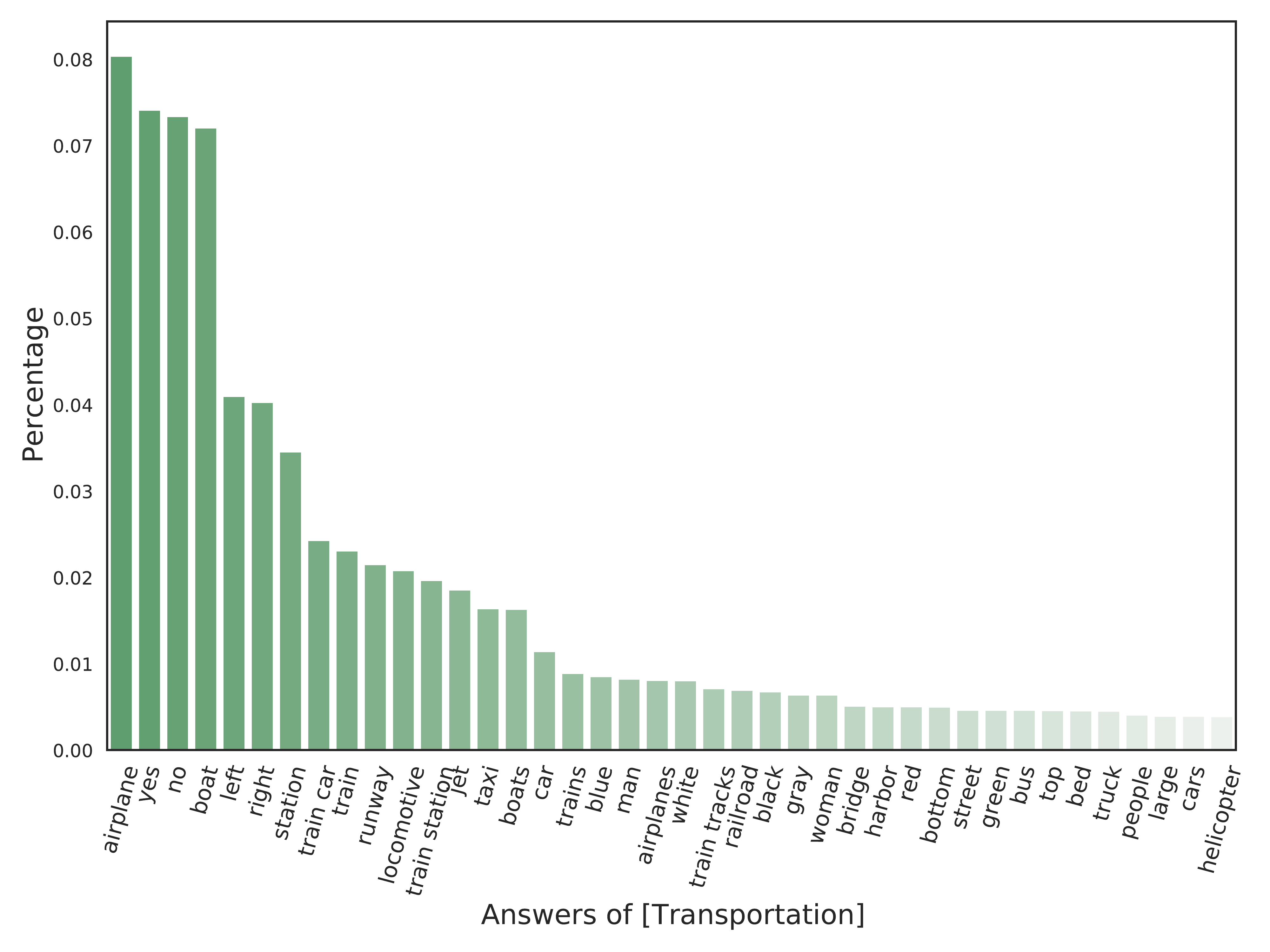} \\
	\includegraphics[width=0.45\textwidth]{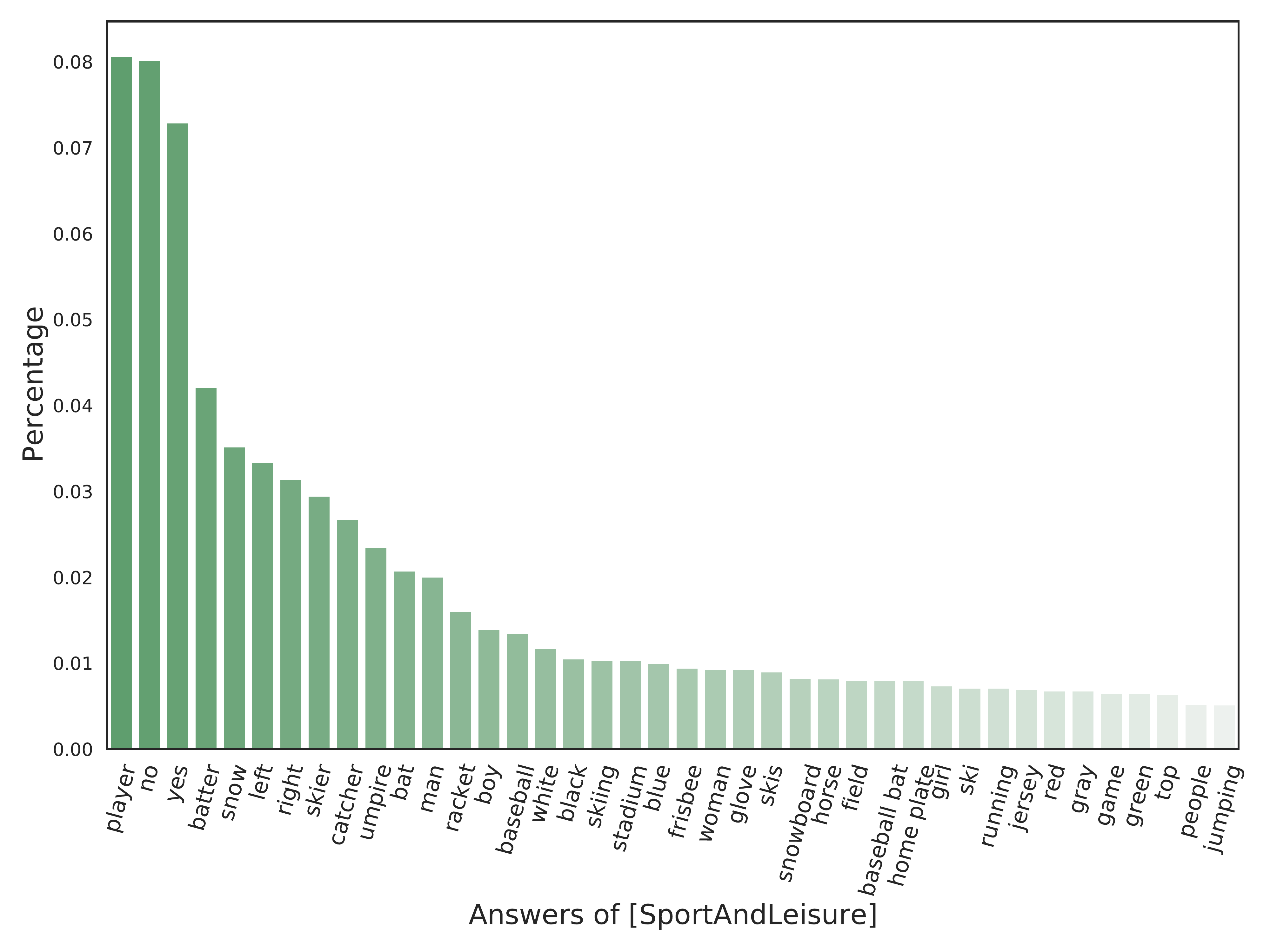}
	\includegraphics[width=0.45\textwidth]{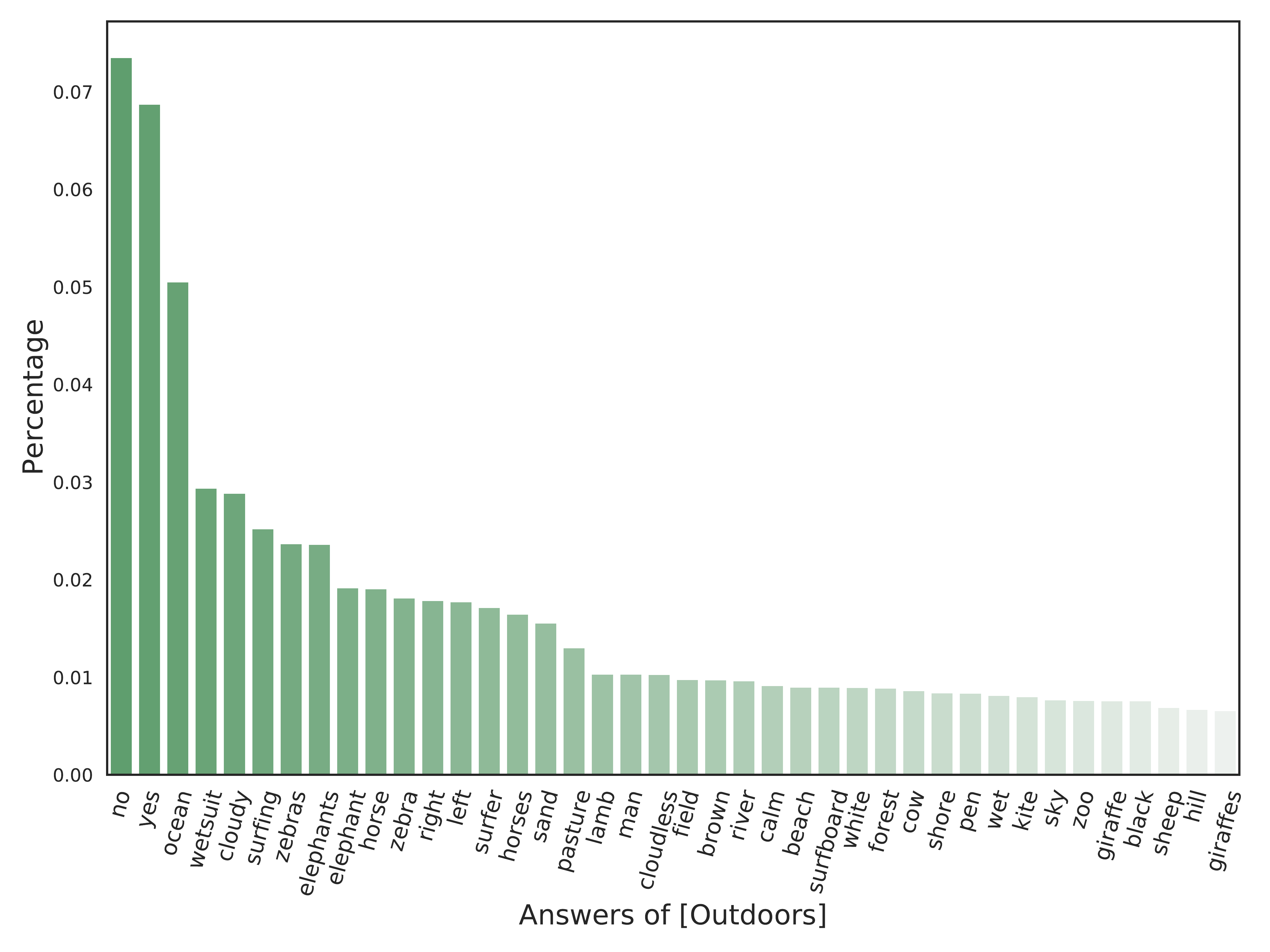}
	
	\caption{Answer distribution of each task in CLOVE-scene. We show the top-40 frequent answers.}
	\label{fig:scene_ans_dist}
\end{figure*}

\begin{figure*}[ht]
\centering
	\includegraphics[width=0.45\textwidth]{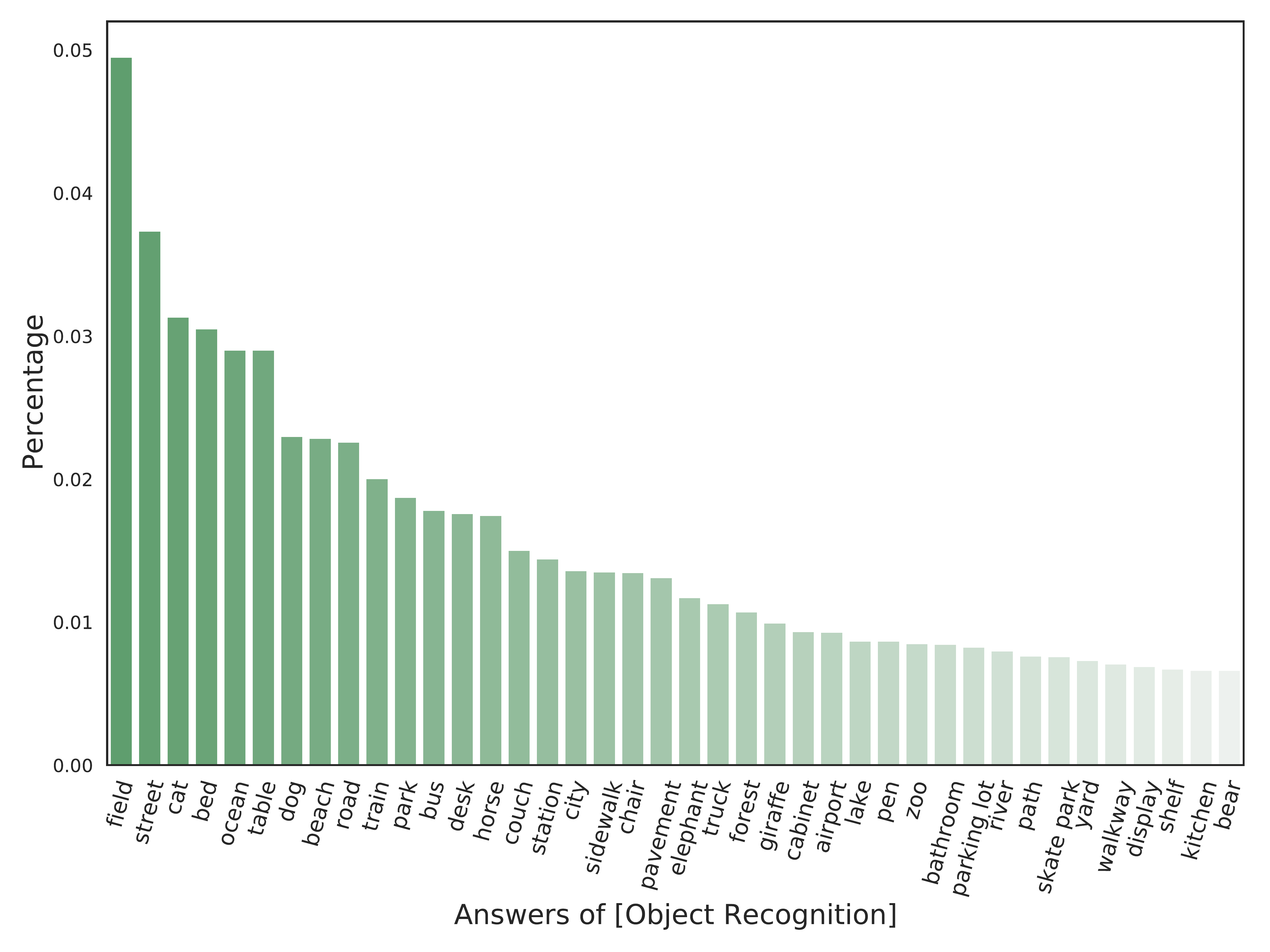}
	\includegraphics[width=0.45\textwidth]{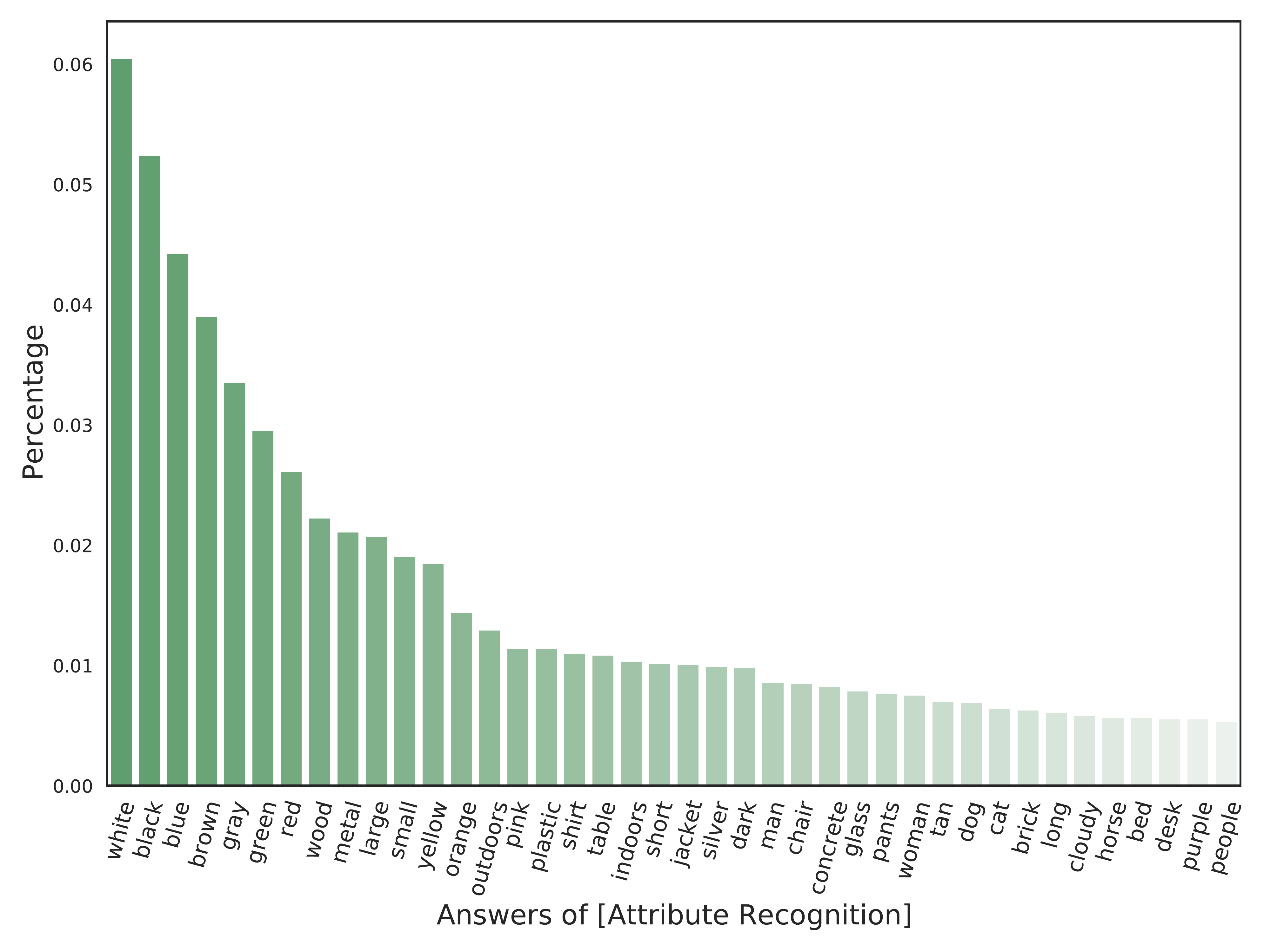} \\
	\includegraphics[width=0.45\textwidth]{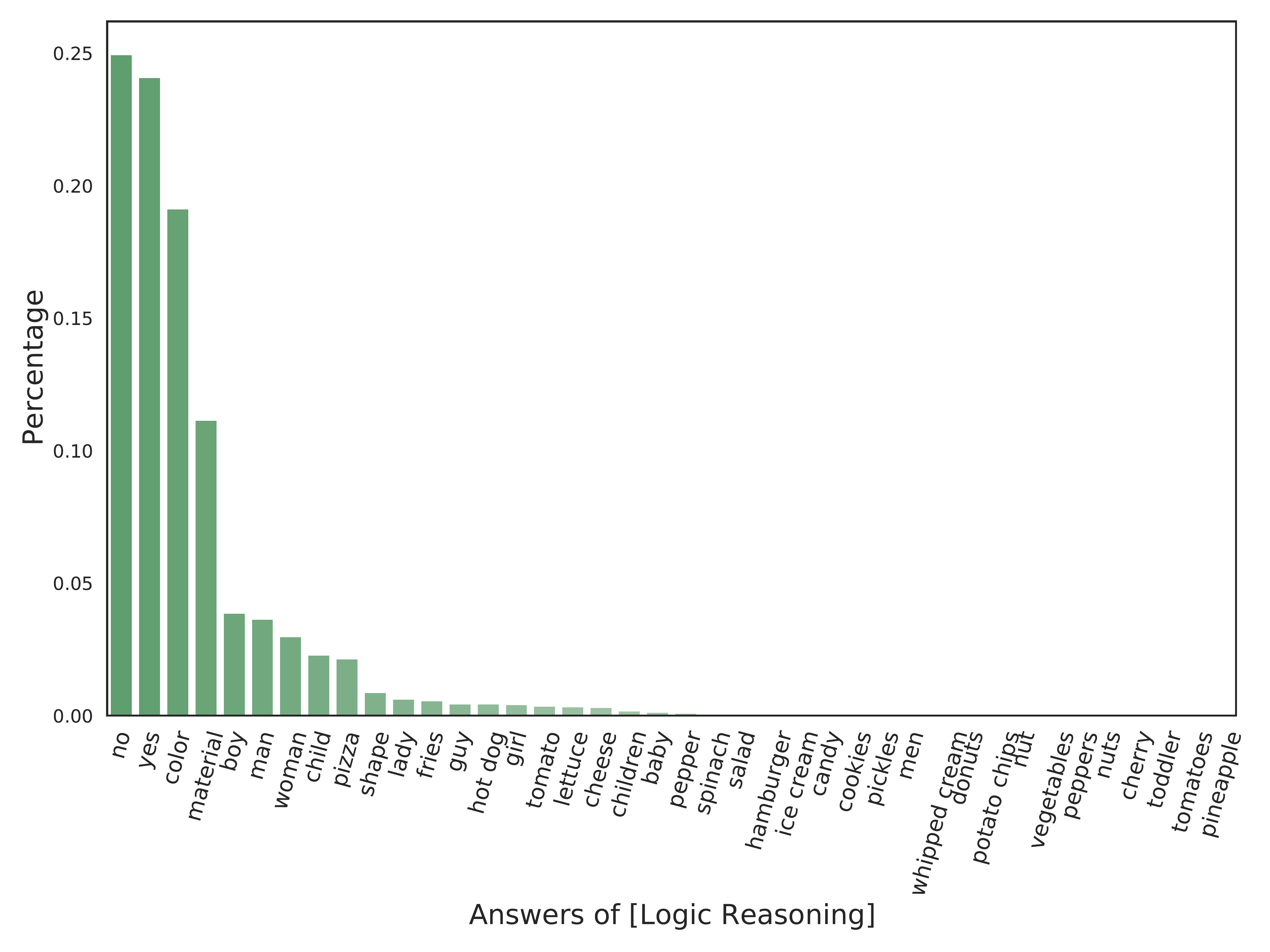}
	\includegraphics[width=0.45\textwidth]{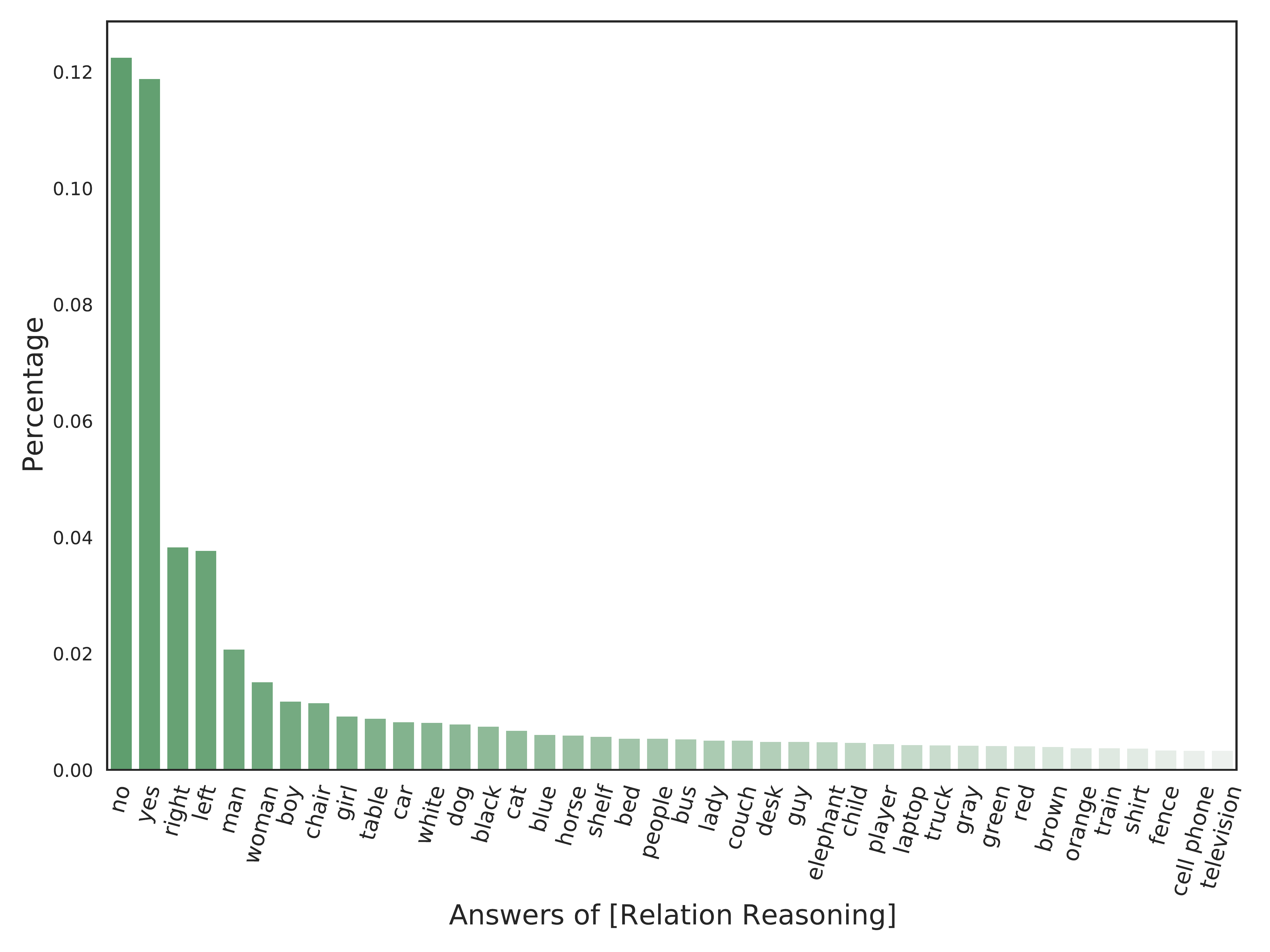} \\
	\includegraphics[width=0.45\textwidth]{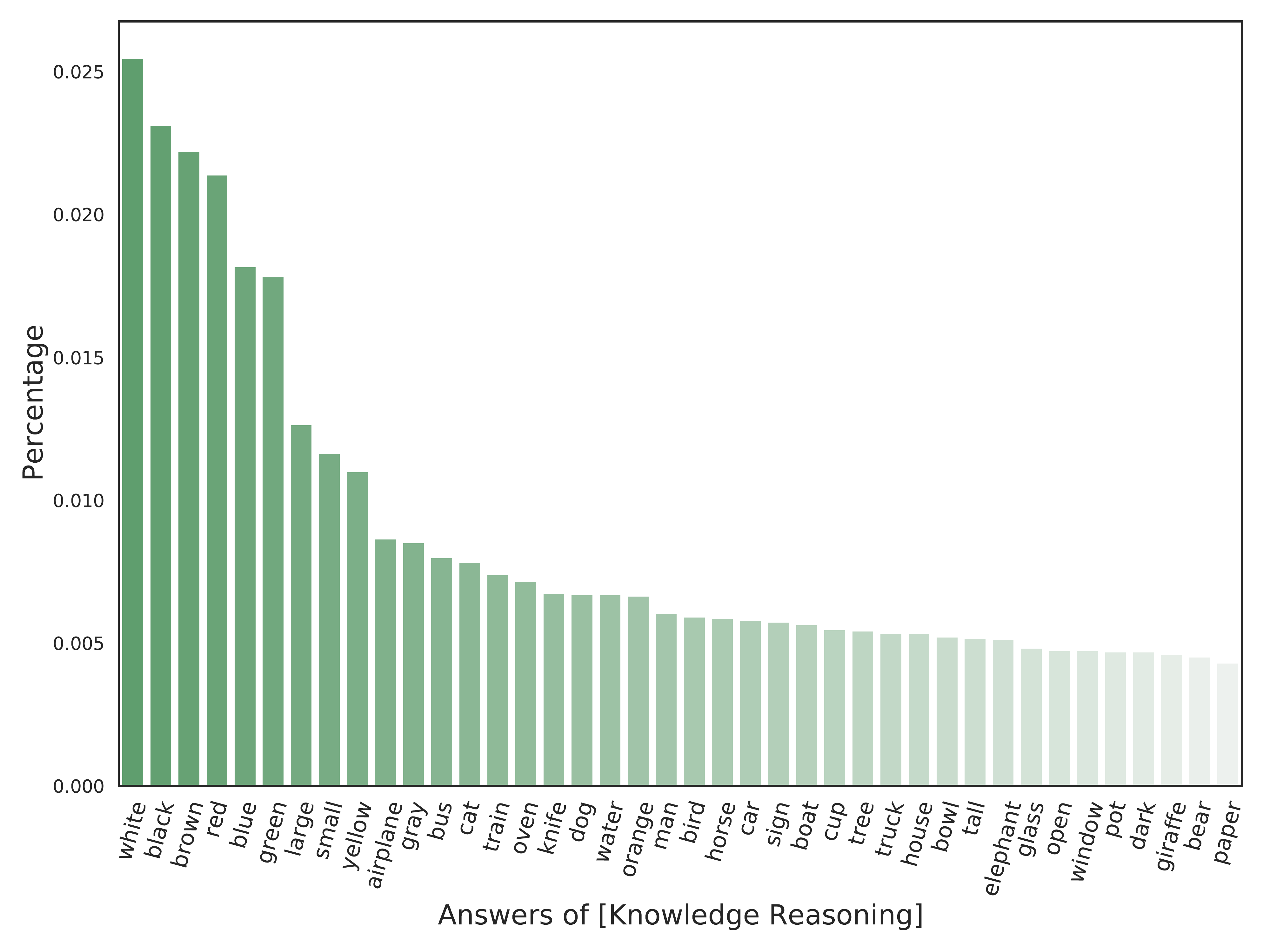}
	\includegraphics[width=0.45\textwidth]{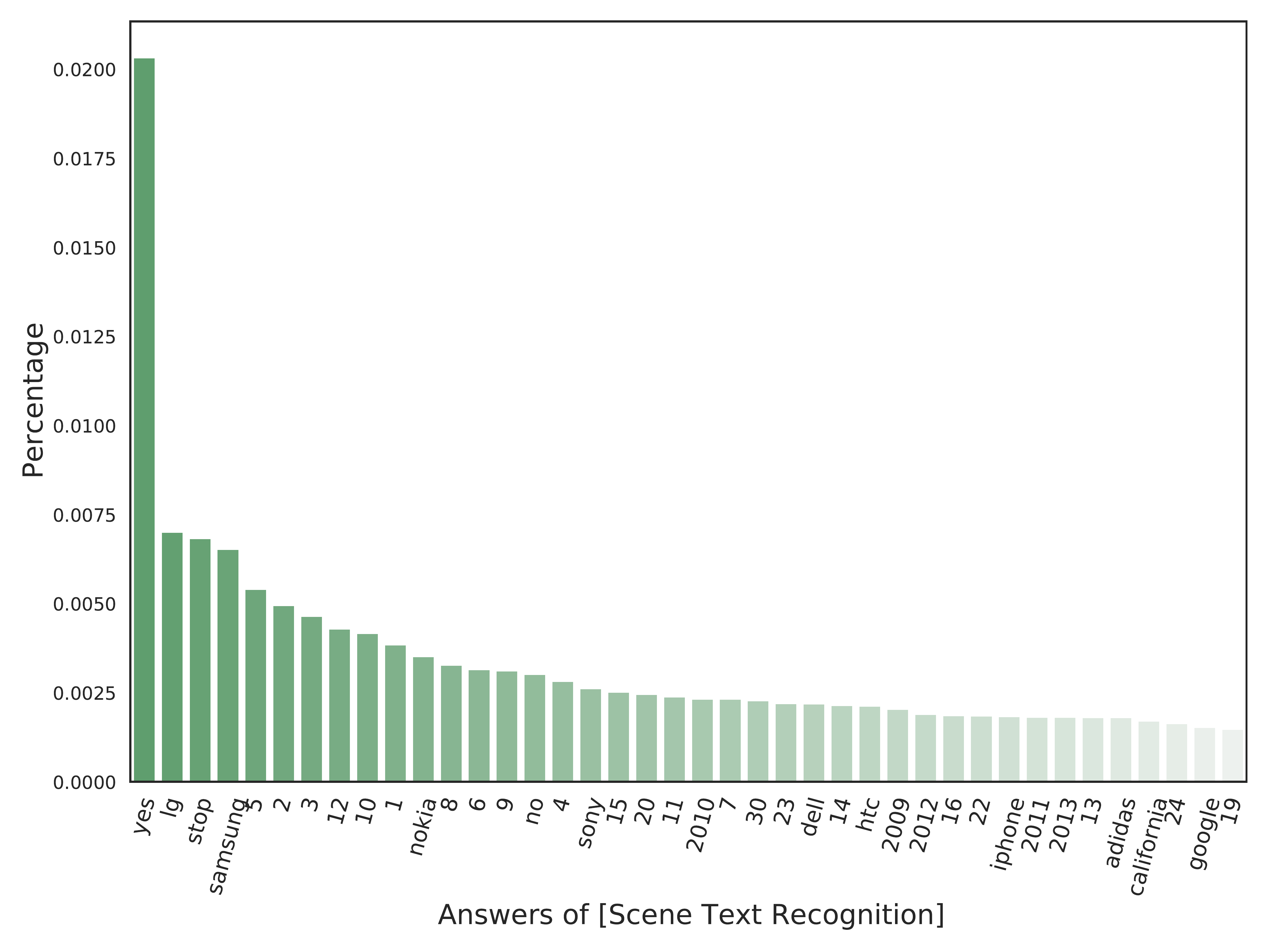}
	
	\caption{Answer distribution of each task in CLOVE-function. We show the top-40 frequent answers.}
	\label{fig:function_ans_dist}
\end{figure*}

We show the answer distribution of the CLOVE-scene and CLOVE-function in~\cref{fig:scene_ans_dist} and~\cref{fig:function_ans_dist}, respectively.

\subsection{More Evaluation Metrics.}
Here, we introduce more metrics~\cite{chaudhry2018riemannian,lopez2017gradient} to measure the performance on CLVQA, which will be used in further experiments:

\head{Average accuracy (A).} Let $a_{k,j}$ denote the accuracy evaluated on the held-out testset of $T_j$($j \leq k$) after training a continual learner from $T_1$ to $T_k$. The average accuracy at $T_k$ is defined as $A_k = \frac{1}{k}\sum_{j=1}^{k}a_{k,j}$.

\head{Forgetting (F).} It is defined as the difference between maximum knowledge gained about the task throughout the learning process in the past and the knowledge that is currently still held about it. The forgetting for $T_j$ after the model has been trained up to $T_k$($k > j$) is:
\begin{equation*}
    f^k_j = \max_{l \in \{1,\cdots,k-1\}} a_{l,j} - a_{k,j}, \forall j<k.
\end{equation*}
Finally, the average forgetting at $T_k$ is $F_k=\frac{1}{k-1}\sum_{j=1}^{k-1} f_j^k$.

\head{Backward transfer (B).} It captures the improvement or deterioration an already observed task experiences when learning a new task. The average backward transfer after training $T_k$ is defined as: $B_k = \frac{1}{k-1}\sum_{j=1}^{k-1}a_{k,j}-a_{j,j}$.

\section{Additional Experimental Results}
\head{Results on different task orders.} Here, we report the Average Accuracy (A), Average Forgetting (F), Average Backward Transfer (B) for six task orders in CLOVE-scene and CLOVE-function, respectively. We use the same task order notation as in the main paper: \textit{oarlks} denotes \textit{object recognition} $\rightarrow$ \textit{attribute recognition }$\rightarrow$ \textit{relation reasoning} $\rightarrow$ \textit{logic reasoning }$\rightarrow$ \textit{knowledge reasoning }$\rightarrow$ \textit{scene text recognition} and \textit{abcdef} denotes \textit{ShopAndDining} $\rightarrow$ \textit{Workplace} $\rightarrow$ \textit{HomeorHotel} $\rightarrow$ \textit{Transportation} $\rightarrow$ \textit{SportAndLeisure} $\rightarrow$ \textit{Outdoors}. For both settings, we set $\gamma=1.5$. Results are shown in~\cref{tab:suppl_results_scene} for CLOVE-scene and~\cref{tab:suppl_results_func} for CLOVE-function.

We can observe that in terms of Average Accuracy (A), Averaging Forgetting (F) and Average Backward Transfer (B), our SGP outperforms all other CL methods without saving real data by a large margin, indicating that our SGP could help better mitigate forgetting to achieve higher accuracy. 
Still, for the three metrics under the CLOVE-scene, our method
underperforms real data replay methods. The reason could be that the disjoint image split in CLOVE-scene increases the difficulty in combating the forgetting in visual modality. Thus, the replayed scene graphs generated by SRM are less informative than images in retaining visual knowledge. For CLOVE-function, our SGP achieves comparable A, F and B when compared to the real data replay methods and even outperforms them in some cases. This is probably because our SRM can capture and remember the reasoning skills for previously seen functions, leading to better replay.

\head{Training process visualization.} We visualize the training progress of six tasks under CLOVE-scene and CLOVE-function, as shown in~\cref{fig:supp_scene_ecdfab} and~\cref{fig:supp_func_oarlks}. The figures record the accuracy on each task as the training proceeds. From the results, we can see that Ours usually performs better on avoiding forgetting, compared to those methods that do not save real samples. Compared with the real data replay method, our approach is slightly inferior on CLOVE-scene because, as we mentioned before, real images really help on tasks involving visual domain increments. On CLOVE-function, our method is comparable to or even better than real data replay, except for the object recognition task. The main reason that ours performs worse on the object recognition task could be that the object recognition questions are less about reasoning and more about the representation of objects, so it is more helpful to retain images than performing symbolic replay.

\head{Examples of generated data.}
~\cref{tab:supp_scene_example} and~\cref{tab:supp_func_example} show the generated scene graphs and QA pairs for each task. It can be seen that our method can generate scene graphs and QA pairs of promising quality. The generated scene graph can make up a reasonable image content. The questions make sense, and the corresponding answers are usually correct.

\begin{table*}[ht]
\begin{tabular}{|p{0.1\textwidth}<{\centering}|p{0.4\textwidth}<{\centering}|p{0.3\textwidth}<{\centering}|p{0.1\textwidth}<{\centering}|}
\hline
Replay   task      & Replay SG                                                                                                                                                                                                                                                                                                                                                                                                                                                                                      & Replay Q                                                           & Replay A \\ \hline
\multirow{2}{*}{a} & \textit{large brownie on   paper white plate {[}S{]}} paper {[}S{]} pizza {[}S{]} sliced red tomato {[}S{]} red tomato on   top of pizza {[}S{]} green pepper {[}S{]} tomato {[}S{]}                                                                                                                                                                                                                                                                                                                    & What dessert is on   the plate?                                    & brownie  \\ \cline{2-4} 
                  & \textit{cake {[}S{]}} sliced red tomato {[}S{]} red tomato on top of white plate {[}S{]}   white plate to the right of sliced red pepper {[}S{]}                                                                                                                                                                                                                                                                                                                                                        & Is the cake in the   top part or in the bottom of the photo?       & top      \\ \hline
\multirow{2}{*}{b} & \textit{silver laptop {[}S{]}} sitting man {[}S{]} man {[}S{]} woman {[}S{]} white shirt {[}S{]} hand {[}S{]}   head {[}S{]} face {[}S{]} hair {[}S{]} shirt {[}S{]} table {[}S{]} pizza {[}S{]} pepperoni {[}S{]} spices   {[}S{]} meat {[}S{]} vegetables {[}S{]} plastic container {[}S{]} bottle {[}S{]} plate {[}S{]}                                                                                                                                                                              & What is the silver   device?                                       & laptop   \\ \cline{2-4} 
                  & \textit{computer to the right of student {[}S{]}} student {[}S{]} man {[}S{]} woman {[}S{]} white   shirt {[}S{]} hand {[}S{]} head {[}S{]} face {[}S{]} hair {[}S{]} shirt {[}S{]} table {[}S{]} pizza {[}S{]}   sliced red tomato {[}S{]} red tomato on top of pizza {[}S{]} small white bowl {[}S{]}   large white round pizza with cheese {[}S{]} cheese {[}S{]} pepperoni {[}S{]}                                                                                                                  & What device is to the   right of the student?                      & computer \\ \hline
\multirow{2}{*}{c} & \textit{pillow {[}S{]}} indoors {[}S{]} wall {[}S{]} bed {[}S{]} lamp {[}S{]} chair {[}S{]} table {[}S{]} pizza   {[}S{]} sliced red tomato {[}S{]} red tomato on top of pizza {[}S{]} small white bowl {[}S{]}   large white round pizza with cheese {[}S{]} cheese {[}S{]} pepperoni {[}S{]}                                                                                                                                                                                                          & Is the pillow on the   right side or on the left?                  & left     \\ \cline{2-4} 
                  & \textit{magnets on refrigerator {[}S{]}} refrigerator to the left of white plate {[}S{]}   white plate on table {[}S{]} table {[}S{]} pizza {[}S{]} sliced red tomato {[}S{]} red tomato   on pizza {[}S{]} green pepper {[}S{]} tomato {[}S{]} pepper {[}S{]} spices {[}S{]} vegetables {[}S{]}   plastic container {[}S{]} bottle {[}S{]}                                                                                                                                                             & What is on the   refrigerator?                                     & magnets  \\ \hline
\multirow{2}{*}{d} & \textit{boat to the right of   boat {[}S{]}} boat {[}S{]} white clouds {[}S{]} blue sky {[}S{]} green leaves {[}S{]} red flower   {[}S{]} yellow flower on top of brown wood table {[}S{]} brown wood chair {[}S{]} small   white bowl {[}S{]} large white round pizza {[}S{]} pizza {[}S{]} sliced red tomato {[}S{]}   tomato {[}S{]} pepperoni {[}S{]} spices {[}S{]} vegetables {[}S{]} plastic container {[}S{]}   bottle {[}S{]}                                                                  & What kind of   watercraft is to the right of the boat on the left? & boat     \\ \cline{2-4} 
                  & \textit{airplane {[}S{]}} outdoors {[}S{]} wing {[}S{]} tail {[}S{]} wheel {[}S{]} wheels {[}S{]}   windows {[}S{]} door {[}S{]} cockpit {[}S{]} trees {[}S{]} windshield {[}S{]} propeller {[}S{]} pilot   {[}S{]} man {[}S{]} woman {[}S{]} person {[}S{]} hand {[}S{]} head {[}S{]} face {[}S{]} eye {[}S{]} ear {[}S{]}                                                                                                                                                                             & What is the aircraft   in the image?                               & airplane \\ \hline
\multirow{2}{*}{e} & \textit{player swinging wood   bat {[}S{]} }wood bat to the left of player {[}S{]} player {[}S{]} white shirt {[}S{]} blue   shirt to beige carpet {[}S{]} green grass {[}S{]} baseball field {[}S{]} gray pants {[}S{]}   brown dirt {[}S{]} man {[}S{]} woman {[}S{]} person {[}S{]} hand {[}S{]} head {[}S{]} face {[}S{]} eye   {[}S{]} ear {[}S{]} mouth {[}S{]} hair {[}S{]} sitting man {[}S{]} standing man wearing glasses   {[}S{]} glasses {[}S{]} glass {[}S{]} bottle {[}S{]} wine {[}S{]} & Who is swinging the   bat?                                         & player   \\ \cline{2-4} 
                  & \textit{skier wearing jacket {[}S{]}} jacket to the left of skier {[}S{]} black pants   {[}S{]} skis {[}S{]} white snow {[}S{]} skiing man {[}S{]} snowy mountain {[}S{]} blue sky {[}S{]}   green leaves {[}S{]} red coat {[}S{]}                                                                                                                                                                                                                                                                      & Who is wearing a   jacket?                                         & skier    \\ \hline
\end{tabular}
\caption{Generated samples of CLOVE-scene at the last task of \textit{abcdef}. SG-prompt is shown in italics.}
\label{tab:supp_scene_example}
\end{table*}

\begin{table*}[ht]
\begin{tabular}{|p{0.1\textwidth}<{\centering}|p{0.4\textwidth}<{\centering}|p{0.3\textwidth}<{\centering}|p{0.1\textwidth}<{\centering}|}
\hline
Replay   task                            & Replay SG                                                                                                                                                                                                                                                        & Replay Q                                                             & Replay A \\ \hline
\multirow{5}{*}{o}    & \textit{horse {[}S{]}} ear {[}S{]}   nose {[}S{]} eye {[}S{]} leg {[}S{]} mouth {[}S{]} head {[}S{]} tail {[}S{]} paw {[}S{]} eyes {[}S{]} face   {[}S{]}                                                                                                                 & What animal is it?                                                   & horse    \\ \cline{2-4} 
                                         & \textit{table {[}S{]}} plate {[}S{]} sauce {[}S{]}                                                                                                                                                                                                                        & What kind of   furniture is it?                                      & table    \\ \cline{2-4} 
                                         & \textit{paddle {[}S{]}} napkin {[}S{]} plate {[}S{]}                                                                                                                                                                                                                      & What is the name of   this animal?                                   & paddle   \\ \cline{2-4} 
                                         & \textit{orange {[}S{]}} napkin {[}S{]} plate {[}S{]}                                                                                                                                                                                                                      & What fruit is it?                                                    & orange   \\ \cline{2-4} 
                                         & \textit{bush {[}S{]}} trees {[}S{]} tree {[}S{]}                                                                                                                                                                                                                          & Which place is it?                                                   & bush     \\ \hline
\multirow{4}{*}{a} & \textit{round clock {[}S{]}}   napkin {[}S{]} sauce {[}S{]} cheese {[}S{]} table {[}S{]}                                                                                                                                                                                  & What shape is the   clock that is on the left?                       & round    \\ \cline{2-4} 
                                         & \textit{black monitor {[}S{]}} keyboard {[}S{]} screen {[}S{]} wall {[}S{]} cord {[}S{]} laptop {[}S{]}                                                                                                                                                                   & What device is black?                                                & monitor  \\ \cline{2-4} 
                                         & \textit{green bus {[}S{]}} window {[}S{]} building {[}S{]} sign {[}S{]} pole {[}S{]}                                                                                                                                                                                      & What color is the bus   on the left?                                 & green    \\ \cline{2-4} 
                                         & \textit{talking woman {[}S{]}} shirt {[}S{]} man {[}S{]} shorts {[}S{]} hair {[}S{]} hand {[}S{]} pants   {[}S{]} person {[}S{]}                                                                                                                                          & Who is talking?                                                      & woman    \\ \hline
\multirow{4}{*}{r}    & \textit{spectator holding   camera {[}S{]} boy to the left of camera {[}S{]}} camera {[}S{]} white shirt {[}S{]} black   pants {[}S{]} hand {[}S{]} head {[}S{]} arm {[}S{]} eye {[}S{]} nose {[}S{]} ear {[}S{]} leg {[}S{]} mouth   {[}S{]} neck {[}S{]} ground {[}S{]} & Who is holding the   device the spectator is to the left of?         & boy      \\ \cline{2-4} 
                                         & \textit{basket full of crusty bread {[}S{]} basket filled with carrot {[}S{]}} carrot on   plate {[}S{]} sauce {[}S{]} cheese {[}S{]} table {[}S{]} plate {[}S{]} pepper {[}S{]}                                                                                          & What is the basket   full of?                                        & bread    \\ \cline{2-4} 
                                         & \textit{white bench in front of wall {[}S{]} bed to the left of wall {[}S{]}} wall {[}S{]}   floor {[}S{]} cabinet {[}S{]} pillow {[}S{]} head {[}S{]} eye {[}S{]} nose {[}S{]} ear {[}S{]} leg {[}S{]}   mouth {[}S{]} neck {[}S{]} ground {[}S{]}                       & What is in front of   the wall the bed is to the left of?            & bench    \\ \cline{2-4} 
                                         & \textit{surfing woman standing in blue water {[}S{]} swimming person near surfing   woman {[}S{]} }                                                                                                                                                                       & Where is the person near the woman standing?                         & water    \\ \hline
\multirow{4}{*}{l}      & \textit{orange beak {[}S{]}   orange pepper {[}S{]} }brown wood table {[}S{]} white napkin {[}S{]} sauce {[}S{]} table {[}S{]}   plate {[}S{]} pepper {[}S{]} crust {[}S{]}                                                                                               & What do both the beak   and the pepper have in common?               & color    \\ \cline{2-4} 
                                         & \textit{green roof {[}S{]} green cap {[}S{]}} white shirt {[}S{]} man {[}S{]} black pants {[}S{]}   hand {[}S{]} head {[}S{]} arm {[}S{]} face {[}S{]} brown hair {[}S{]} blue jeans {[}S{]} woman {[}S{]}   girl {[}S{]} boy {[}S{]} person {[}S{]}                      & What do both the cap   and the roof have in common?                  & color    \\ \cline{2-4} 
                                         & \textit{plastic container {[}S{]} plastic bag {[}S{]}} brown wood table {[}S{]} white napkin {[}S{]}   sauce {[}S{]} table {[}S{]} plate {[}S{]} pepper {[}S{]} crust {[}S{]}                                                                                             & What do both the   container and the bag have in common?             & material \\ \cline{2-4} 
                                         & \textit{long window {[}S{]} long tail {[}S{]} }blue clear sky {[}S{]}                                                                                                                                                                                                     & What do the window   and the tail have in common?                    & color    \\ \hline
\multirow{4}{*}{k}   & \textit{floor is used for walking on {[}S{]} red floor {[}S{]}} towel hanging on wall {[}S{]}   handle on door {[}S{]} mirror on window {[}S{]}                                                                                                                           & what color is the   thing that I can use for walking on              & red      \\ \cline{2-4} 
                                         & \textit{window can be opened or closed {[}S{]} bright window {[}S{]}} window on train   {[}S{]}                                                                                                                                                                           & is the thing that can   be opened or closed bright or dark           & bright   \\ \cline{2-4} 
                                         & \textit{table is used for eating at it {[}S{]} black table {[}S{]}} man wearing shirt   {[}S{]} woman has hair {[}S{]} hand holding phone {[}S{]}                                                                                                                         & what color is the   object that I can use for eating at it           & black    \\ \cline{2-4} 
                                         & \textit{field is used for grazing animals {[}S{]} open field {[}S{]}} man wearing hat   {[}S{]} hat on head {[}S{]}                                                                                                                                                       & is the place that   could be used for grazing animals open or closed & open     \\ \hline
\end{tabular}
\caption{Generated samples of CLOVE-function at the last task of \textit{oarlks}. SG-pormpt is shown in italics.}
\label{tab:supp_func_example}
\end{table*}

\clearpage

\clearpage

{\small
\bibliographystyle{ieee_fullname}
\bibliography{egbib}
}

\end{document}